\documentclass[9.5pt,final]{IEEEtran}
\usepackage{color,soul}
\usepackage{graphicx}
\usepackage{cite}
\usepackage[cmex10]{amsmath}
\usepackage{amssymb}
\usepackage{subfigure}
\usepackage{rotating}
\usepackage{multirow}
\usepackage{makecell}
\usepackage[colorlinks=true,linkcolor=red,citecolor=blue]{hyperref}
\usepackage{stfloats}
\usepackage{caption}
\usepackage{balance}
\usepackage{mathrsfs}
\usepackage{algorithm}
\usepackage{booktabs}
\usepackage{multicol}
\usepackage{float}
\usepackage{array}
\usepackage{colortbl}
\usepackage{siunitx}
\usepackage[keeplastbox]{flushend}
\usepackage{threeparttable}
\newcounter{magicrownumbers}  % number in table

\usepackage[utf8]{inputenc}
\usepackage{ragged2e}
\usepackage{soul}
\usepackage{bbding}
\usepackage{pifont}

\def\eg{{\em e.g.}} 

\def\ie{{\em i.e.}}

\soulregister\cite7 
\soulregister\citep7 
\soulregister\citet7 
\soulregister\ref7
\soulregister\eg7
\soulregister\quad7

\begin{document}

\title{On Creating Benchmark Dataset for Aerial Image Interpretation: Reviews, Guidances and Million-AID}


\author{Yang Long, 
\and Gui-Song~Xia,~\IEEEmembership{Senior Member,~IEEE}, 
\and Shengyang Li, 
\and Wen Yang,~\IEEEmembership{Senior Member,~IEEE},  \\
\and Michael Ying Yang,~\IEEEmembership{Senior Member,~IEEE},  
\and Xiao Xiang Zhu,~\IEEEmembership{Fellow,~IEEE},  
\\
\and Liangpei Zhang,~\IEEEmembership{Fellow,~IEEE}, 
\and Deren Li 
\vspace{-2mm}
\IEEEcompsocitemizethanks{
  \IEEEcompsocthanksitem{
%\thanks{
The study of this paper is funded by the National Natural Science Foundation of China (NSFC) under grant contracts No.61922065, No.61771350 and No.41820104006 and 61871299. It is also funded by the Science and Technology Major Project of Hubei Province (Next-Generation AI Technologies) under Grant 2019AEA170. It is also partially supported by the German Federal Ministry of Education and Research (BMBF) in the framework of the international future AI lab ``AI4EO -- Artificial Intelligence for Earth Observation: Reasoning, Uncertainties, Ethics and Beyond''. } % <-this % stops a space
\IEEEcompsocthanksitem{
%\thanks{
Y. Long, L. Zhang, D. Li are with the State Key Lab. LIESMARS, Wuhan University, Wuhan, China. 
e-mail: \{{\em longyang, zlp62, drli}\}@whu.edu.cn}% <-this % stops a space
\IEEEcompsocthanksitem{
%\thanks{
G.-S. Xia is with the School of Computer Science and also the State Key Lab. LIESMARS, Wuhan University, Wuhan, China. 
e-mail: guisong.xia@whu.edu.cn}% <-this % stops a space
\IEEEcompsocthanksitem{
%\thanks{
S. Li is with the Key Laboratory of Space Utilization, Technology and Engineering Center for Space Utilization, Chinese Academy of Sciences, Beijing, China. 
e-mail: shyli@csu.ac.cn}% <-this % stops a space
\IEEEcompsocthanksitem{
%\thanks{
W. Yang is with the School of Electronic Information and the State Key Lab. LIESMARS, Wuhan University, Wuhan China.
e-mail: yangwen@whu.edu.cn}% <-this % stops a space
\IEEEcompsocthanksitem{
%\thanks{
M. Y. Yang is with the Faculty of Geo-Information Science and Earth Observation, University of Twente, Hengelosestraat 99, Enschede, Netherlands.
e-mail: michael.yang@utwente.nl}% <-this % stops a space
\IEEEcompsocthanksitem{
%\thanks{
X. Zhu is with the German Aerospace Center (DLR) and also Technical University of Munich, Germany.
e-mail: xiaoxiang.zhu@dlr.de}% <-this % stops a space
\IEEEcompsocthanksitem{
%\thanks{
Corresponding author: Gui-Song Xia (guisong.xia@whu.edu.cn).}
}
}

% \vskip -2.1\baselineskip plus -1fil

\IEEEtitleabstractindextext{
\begin{abstract}
\justifying
The past years have witnessed great progress on remote sensing (RS) image interpretation and its wide applications. With RS images becoming more accessible than ever before, there is an increasing demand for the automatic interpretation of these images. In this context, the benchmark datasets serve as essential prerequisites for developing and testing intelligent interpretation algorithms. After reviewing existing benchmark datasets in the research community of RS image interpretation, this article discusses the problem of how to efficiently prepare a suitable benchmark dataset for RS image interpretation. Specifically, we first analyze the current challenges of developing intelligent algorithms for RS image interpretation with bibliometric investigations. We then present the general guidances on creating benchmark datasets in efficient manners. Following the presented guidances, we also provide an example on building RS image dataset, \ie, \textbf{Million-AID}~\footnote{A website is available at:~\href{https://captain-whu.github.io/DiRS/}{\textcolor[RGB]{236,0,140}{https://captain-whu.github.io/DiRS/}}}, a new large-scale benchmark dataset containing a million instances for RS image scene classification. Several challenges and perspectives in RS image annotation are finally discussed to facilitate the research in benchmark dataset construction. We do hope this paper will provide the RS community an overall perspective on constructing large-scale and practical image datasets for further research, especially data-driven ones.
\end{abstract}
%~\footnote{A website is available at:~\href{https://captain-whu.github.io/DiRS/}{\textcolor[RGB]{236,0,140}{https://captain-whu.github.io/DiRS/}}}
\begin{IEEEkeywords}
Remote sensing image interpretation, annotation, benchmark datasets, scene classification, Million-AID
\end{IEEEkeywords}}

\maketitle
\IEEEdisplaynontitleabstractindextext

\section{Introduction} \label{Introduction}
\IEEEPARstart{T}{he} advancement of remote sensing (RS) technology has significantly improved the ability of human beings to characterize features of the Earth surface~\cite{Toth201622,xiang2019mini}. With more and more RS images being available, the interpretation of RS images has been playing an important role in many applications, such as environmental monitoring~\cite{ZHENG20191,BAUER2020111522}, resource investigation~\cite{Tong2019GID,SHAKER201994,SHENDRYK2019124}, and urban planning~\cite{coutts2016thermal,ZHOU2020111458}, {\em etc}. However, with the rapid development of the Earth observation technology, the volume of RS images increases dramatically, which raises high requirement of efficient image interpretation for real-world applications. Moreover, the rich details in RS images, such as the geometrical shapes, structural characteristics, and textural attributes also pose great challenges to the interpretation of image content~\cite{xia2010structural,xia2017texture,ANWER201874,HSFNET2018}. These motivate the increasing and stringent demands for automatic and intelligent interpretation of the blooming RS imagery.

To characterize RS image content, quite a few methods have been developed for various interpretation tasks, ranging from the scene-level content recognition~\cite{UCM,sheng2012high,Yang2015Scene,DeepMetric2018,tong2019exploiting,ScaleFreeSC2019,Zheng2019SC,liang2020deep,bi2020multiple,cheng2020deepsc}, object-level image analysis~\cite{porway2010hierarchical,han2014object,cheng2016survey,zhang2016weakly,DOTA,ding2019learning,zou2019object,ORSIm2019Detector,hossain2019obia,Ma2017277,Gliding2020} to the challenging pixel-wise semantic understanding~\cite{tuia2009active,tuia2011survey,romero2015unsupervised,maggiori2016convolutional,KernelSS,Zhu2017370,zhang2018coarse,Change2019,GEI201942,MultilevelCD,hong2020graph,zhang2020deep}. Benefiting from the increasing availability and various ontologies of RS images, the developed methods have reported promising performance on the interpretation of RS image content. However, many of the current methods are evaluated on small-scale image datasets which usually show domain bias for applications. Moreover, a dataset created toward specific algorithms rather than real application scenarios is hard to objectively validate the comprehensive performance of the algorithms. Recently, it is observed that data-driven approaches, particularly the deep learning ones~\cite{deepreview,ghamisi2017advanced,reichstein2019deep,hong2020more}, have become an important alternative to manual interpretation and provided a bright prospect for automatic interpretation, analysis and content understanding for the massive RS images. However, the training and testing effectiveness could be curbed owing to the lack of adequate and accurately annotated ground-truth datasets. As a result, it usually turns out to be difficult to apply the interpretation models in real-world applications. Thus, it is natural to argue that a great amount of efforts need to be paid for datasets construction considering the following points:

 \begin{itemize}
   \item 
   {\em The ever-growing volume of RS images is acquired while very few of them are annotated with valuable information.} With the rapid development and continuous improvement of sensor technology, it is convenient to receive RS data with various modalities, \eg, optical, hyper-spectral, and synthetic aperture radar (SAR) images. Consequently, a huge amount of RS images with different spatial, spectral, and temporal resolutions is received every day than ever before, providing challenges as well as opportunities~\cite{7565634} for the interpretation of surface features~\cite{AID,DOTA,Tong2019GID}. However, in contrast to the huge amount of received RS images, those annotated with valuable information are relatively few, making them difficult to be productively utilized and also resulting in great waste.

   \item
   {\em The generalization ability of algorithms for interpreting RS images is of great urgency to be enhanced.} Although a multitude of machine learning~\cite{mountrakis2011support,belgiu2016random,LAGRANGE201926} and deep learning algorithms~\cite{hu2015transferring,deepreview,ma2019deep} have been developed for RS image interpretation, their interpretation capability could be constrained because of the complexity of RS image content. Besides, existing algorithms are usually trained on small-scale datasets, which shows weak representation ability for the real-world feature distribution. Consequently, the constructed algorithms inevitably show limitations, \eg, weak generalization ability, in practical applications. Therefore, more robust and intelligent algorithms need to be further explored accounting for the essential characteristics of RS images. 

   \item
   {\em Representative and large-scale RS image datasets with accurate annotations are demanded to narrow the gap between algorithm development and real applications.} An annotated dataset with large volume and variety has proven to be crucial for feature learning~\cite{Imagenet,PASCALVOC,COCO,DOTA}. Although various datasets have been built for different RS image interpretation tasks, there are inadequacies, \eg, the small scale of images, the limited semantic categories, and deficiencies in image diversity, which severely limit the development of new approaches. From another point of view, large-scale datasets are more conducive to characterize the pattern of feature distribution in the real-world. Thus, it is natural to argue that the representative and large-scale RS image datasets are critical to push forward the development of practical interpretation algorithms, particularly deep learning-based methods. 

   \item 
   {\em There is a lack of public platforms for systematic evaluation and fair comparison among different interpretation algorithms.} A host of interpretation algorithms have been designed for RS image interpretation tasks and achieved excellent performances. However, many algorithms are designed toward specific datasets, rather than practical applications. Without the persuasive evaluation and comparison platforms, it is an arduous task to fairly compare and optimize different algorithms. Moreover, the established image datasets may show deficiencies in scale, diversity and other properties as mentioned before. This makes the learned algorithms inherently deficient. As a result, it is difficult to effectively and systematically measure the validity and practicability of different algorithms for real interpretation applications. 
 \end{itemize}

With these points in mind, this paper first provides a review of the available RS image datasets and discusses the creation of benchmark datasets for RS image interpretation. Then, we present an example of constructing a large-scale dataset for scene classification as well as the discussion about challenges and perspectives in RS image annotation. To sum up, our main contributions are as follows: 
 \begin{itemize}
     \item 
     Covering literature published over the past decade, we provide a comprehensive review on the existing RS image datasets concerning the current mainstream of RS image interpretation tasks, including scene classification, object detection, semantic segmentation, and change detection. 
     
     \item We present the general guidances, including the dataset property desirability, image acquisition via semantic coordinates collection, and annotation methodology, on creating benchmark datasets for RS image interpretation. The introduced guidances formulate an overall prototype, which we hope to provide a picture for RS image dataset creation with considerations in efficiency, quality assurance, and property assessment.
    
    \item Following the general guidances of dataset creation, we establish the solution of building a scene classification dataset to further verify the practicability of the formed guidances. Consequently, we create a large-scale benchmark dataset for RS image scene classification, \ie, \textbf{Million-AID}, which possesses a million RS images. Besides, we conduct a discussion about the challenges and perspectives in RS image dataset annotation to which efforts need to be dedicated in the future work.
 \end{itemize}

 The remainder of this paper is organized as follows. Section~\ref{DatasetReview} reviews the existing datasets for RS image interpretation. Section~\ref{Conception} presents the guidances of constructing a meaningful annotated RS image dataset. Section~\ref{ExampleMAID} gives an example of creating large-scale RS image dataset for scene classification. Section~\ref{Resources} discusses the challenges and perspectives concerning RS image annotation. Finally, in Section~\ref{Conclusions}, we draw some conclusions.

\section{Annotated Datasets for RS Image Interpretation: A Review } \label{DatasetReview}
The interpretation of RS images has been playing an increasingly important role in a large variety of applications, and thus, has attracted remarkable research attentions. Consequently, many RS image datasets have been built to advance the development of interpretation algorithms. In this section, we firstly investigate the mainstream of RS image interpretation. And then, a comprehensive review is conducted from the perspective of dataset annotation. 

\subsection{RS Image Interpretation Focus in the Past Decade}
It is of great interest to check what the main research stream is in RS image interpretation. To do so, we analyzed the journal articles published in the past decade in RS community based on Web of Science (WoS) database. Specifically, we use ``remote sensing'' as the keyword to perform {\em topic} retrieval supported by {\em tile}, {\em abstract}, and {\em keywords}. Then, the retrieved references published in the last decade (\ie,~2011-2020) are gathered and those journals that published most articles ranked top 10 are selected to investigate the mainstream of RS image interpretation. Generally, remote sensing image interpretation is closely related to the work of image/information/content extract/analysis/understanding. Relying on this idea, each term of ``image interpretation'', ``image analysis'', ``image understanding'', ``content interpretation'', ``content analysis'', ``content understanding'', ``content extraction'', ``information extraction'', ``information analysis'', ``information interpretation'', and ``information understanding'' was combined with the keyword of ``remote sensing'' to further screen those interpretation related works by topic retrieval. By excluding the irrelevant results (\eg, review articles), 5,827 articles were obtained and then analyzed by CiteSpace~\cite{citespace}. Table~\ref{tab:journals} shows the final employed journals and number distribution of investigated references. It is shown that our investigated references are now well presented at the major international RS journals.

\begin{table}[h]
\centering
	\caption{Investigated Journals and number of papers.}\label{tab:journals}
	\vspace{-1mm}
		\begin{tabular}{lc}
			\hline
	            Name of journal &\#Ref. \\
	        \hline
                Remote Sensing  &1,922  \\ 
                International Journal of Remote Sensing &587 \\
                IEEE Transactions on Geoscience and Remote Sensing &575 \\
                ISPRS Journal of Photogrammetry and Remote Sensing &536 \\
                Remote Sensing of Environment &528 \\
                IEEE Journal of Selected Topics in Applied Earth  &493 \\
                \qquad Observations and Remote Sensing \\
                Journal OF Applied Remote Sensing &329 \\
                International Journal of Applied Earth Observation and &304 \\
                \qquad Geoinformation \\
                Sensors &277 \\
                IEEE Geoscience and Remote Sensing Letters &276 \\
			\hline
		\end{tabular}
\end{table}

\begin{figure}[h]
  \centering
  \includegraphics[width=0.99\linewidth]
  {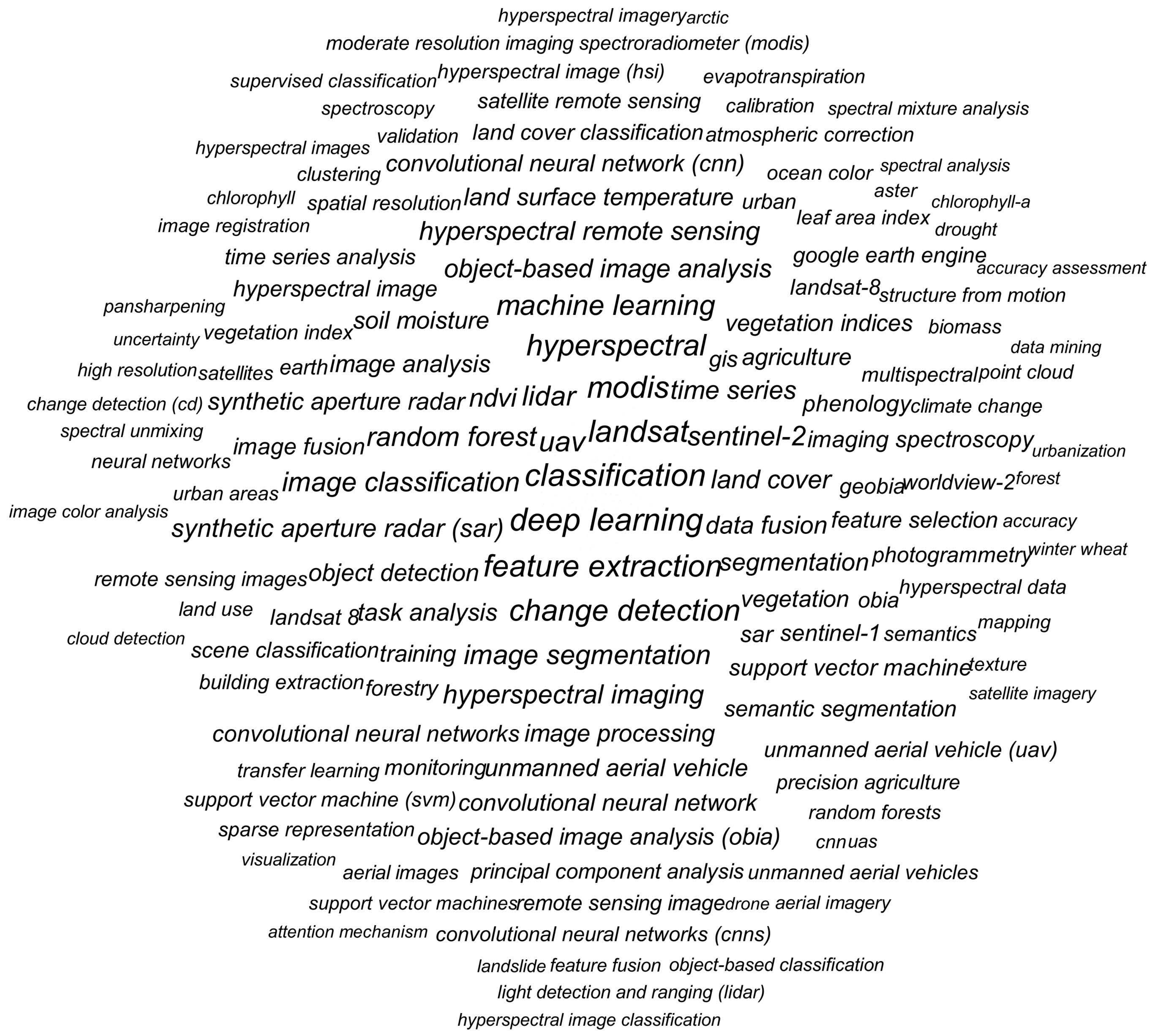}
  \caption{Tag cloud of RS image interpretation.}
  \label{figure:KeyWords}
  \vspace{-3mm}
\end{figure}

 Figure~\ref{figure:KeyWords} shows the highest frequency terms appearing in the title, keyword, and abstract of the literature. The terms with higher frequency are presented with larger font size. As can be seen from this figure, RS image interpretation works mainly focus on {\em classification} tasks (\eg, land-cover classification and scene classification). Obviously, {\em change detection}, {\em (image) segmentation}, and {\em object detection} occupy prominent positions in the interpretation tasks. Specially, the terms around the center, {\eg}, {\em landsat}, {\em uav} (unmaned aerial vehicle), {\em modis}, {\em synthetic aperture radar}, and {\em sentinel}*, indicate the commonly used image sources for interpretation tasks. It is worth noting that {\em feature extraction} plays a significant role in the interpretation of RS images. This makes sense as the feature extraction performed by interpretation models and algorithms, reflected by the terms of {\em deep learning}, {\em machine learning}, {\em convolutional neural network} (CNN), {\em random forest}, and {\em support vector machine}), is indispensable to RS image interpretation tasks. Notably,~{\em deep learning} represented by~{\em convolutional neural network} also occupies the center of the tag cloud, where the currently most popular method for RS image interpretation is revealed. And this has heavily promoted dataset construction to advance the development of RS image interpretation. We subsequently filtered the meta articles by ``deep learning'' and ``convolutional neural network''. The highest-frequency terms match well with Figure~\ref{figure:KeyWords}, where scene classification, object detection, segmentation, and change detection possess the centrality of interpretation tasks, verified by~\cite{ma2019deep}. Thus, the review given below focuses mainly on datasets concerning these topics.
 
\subsection{Annotated Datasets for RS Image Interpretation}
During the past decade, a number of datasets for RS image interpretation have been released publicly. The available datasets are arranged in chronological order as shown in Tables~\ref{tab:SCdataset}-\ref{tab:CDdataset}, in which the corresponding references can be referred for more detailed information about these datasets. Instead of simply delivering descriptions of the datasets, we focus on analyzing the properties of the public RS image datasets from the perspective of annotation~\footnote{We pay our attention mainly to the publicly released and popular RS image datasets while those for special domains,~\eg, contest and private applications, may not be fully covered due to their unstable accessibility or incomplete dataset information.}.

\begin{table*}[!ht]
    \centering
	\caption{Comparison among different RS image scene classification datasets.}\label{tab:SCdataset}
	\vspace{-1mm}
	  \begin{threeparttable}
		\begin{tabular}{lccccccc}
			\hline
	Dataset &\#Cat. &\#Images per cat. &\#Instances &Resolution (m) &Image size &GL/IT/SP &Year\\
	        \hline
    UC-Merced~\cite{UCM}  &21  &100   &2,100 &0.3  &256$\times$256 &\ding{53}\,\ding{53}\,\ding{53} &2010\\ 
    WHU-RS19~\cite{xia2010structural}  &19  &50 to 61  &1,013  &up to 0.5  &600$\times$600 &\ding{53}\,\ding{53}\,\ding{53}  &2012\\
    RSSCN7~\cite{RSSCN7}  &7  &400  &2,800  &--  &400$\times$400 &\ding{53}\,\ding{53}\,\ding{53}  &2015\\
    SAT-4~\cite{SAT-6}  &4  &89,963 to 178,034  &500,000  &1 to 6  &28$\times$28 &\ding{53}\,\ding{53}\,\ding{53} &2015\\
    SAT-6~\cite{SAT-6}  &6  &10,262 to 150,400 &405,000  &1 to 6  &28$\times$28 &\ding{53}\,\ding{53}\,\ding{53}  &2015\\
    BCS~\cite{BSC2015}  &2  &1,438  &2,876  &--  &600$\times$600  &\ding{53}\,\ding{53}\,\checkmark &2015\\
    RSC11~\cite{RSC11}  &11  &$\sim$100  &1,232  &$\sim$0.2  &512$\times$512 &\ding{53}\,\ding{53}\,\ding{53} &2016\\
    SIRI-WHU~\cite{SIRI}  &12  &200  &2,400  &2  &200$\times$200 &\ding{53}\,\ding{53}\,\ding{53}  &2016\\
    NWPU-RESISC45~\cite{RESISC45}  &45  &700  &31,500  &0.2 to 30  &256$\times$256 &\ding{53}\,\ding{53}\,\ding{53}  &2016\\
    AID~\cite{AID}  &30  &220 to 420  &10,000  &0.5 to 8  &600$\times$600 &\ding{53}\,\ding{53}\,\ding{53} &2017\\
    RSI-CB256~\cite{RSI-CB}  &35  &198 to 1,331  &24,000  &0.3 to 3  &256$\times$256 &\ding{53}\,\ding{53}\,\ding{53}  &2017\\
    RSI-CB128~\cite{RSI-CB}  &45  &173 to 1,550  &36,000  &0.3 to 3  &128$\times$128 &\ding{53}\,\ding{53}\,\ding{53}  &2017\\
    Planet-UAS~\cite{Planet-UAS} &17  & --  &40,480  &3 to 5  &256$\times$256 &\checkmark\,\checkmark\,\checkmark &2017\\
    RSD46-WHU~\cite{RSD46-WHU}  &46  &500 to 3,000  &117,000  &0.5 to 2  &256$\times$256 &\ding{53}\,\ding{53}\,\ding{53} &2017\\
    MASATI~\cite{MASATI} &7  & 304 to 1,789  &7,389   & --  &512$\times$512 &\ding{53}\,\ding{53}\,\ding{53} &2018\\
    EuroSAT~\cite{EuroSAT}  &10  &2,000 to 3,000  &27,000  &10  &64$\times$64 &\checkmark\,\checkmark\,\checkmark &2018\\
    PatternNet~\cite{PatternNet}  &38  &800  &30,400  &0.06 to 4.7  &256$\times$256 &\ding{53}\,\ding{53}\,\ding{53}  &2018\\
    fMoW~\cite{fWoM}  &62  &--  &132,716  &~0.5  &74$\times$58 to 16,184$\times$16,288 &\checkmark\,\checkmark\,\checkmark  &2018\\
    WiDS Datathon 2019~\cite{WiDS}  &2  &--  &20,000  &3  &256$\times$256  &\ding{53}\,\ding{53}\,\ding{53} &2019\\
    Optimal-31~\cite{optimal31}  &31  &60  &1,860  &--  &256$\times$256 &\ding{53}\,\ding{53}\,\ding{53} &2019\\
    BigEarthNet~\cite{BigEarthNet}  &43  &328 to 217,119  &590,326  &10,20,60  &20$\times$20;60$\times$60;120$\times$120 &\checkmark\,\checkmark\,\checkmark  &2019\\
    CLRS~\cite{CLRS}  &25  &600  &15,000  &0.26 to 8.85  &256$\times$256 &\ding{53}\,\ding{53}\,\ding{53}  &2020\\
    MLRSN~\cite{MLRSNet} &46  &1,500 to 3,000  &109,161  &0.1 to 10  &256$\times$256 &\ding{53}\,\ding{53}\,\ding{53}  &2020\\
			\hline
		\end{tabular}
		\begin{tablenotes}
    	\scriptsize
    	    \item[*] As~{\em fMoW} is constructed with multiple temporal views for each scene, we ignore the~{\em \#Images per Cat.} and count the total number of unique scene instances,~\ie, \#Instances. Note that~{\em MLRSN} is a multi-label scene classification dataset. The {\em Cat.},~{\em GL}, {\em IT}, and {\em SP} are short for {\em Category},~{\em Geographic Location}, {\em Imaging \;Time}, and {\em Sensor parameter}, respectively. We present the~{\em GL/IT/SP} column to indicate whether the datasets provide those complete and accurate meta information.
    	\end{tablenotes}
	  \end{threeparttable}
\end{table*}

\begin{table*}[!t]
    \centering
	\caption{Comparison among different RS Image object detection datasets.}\label{tab:ODdataset}
	\vspace{-1mm}
	  \begin{threeparttable}
		\begin{tabular}{lcccccccc}
			\hline
            Datasets                          & Annot.  & \#Cat. & \#Instances  &\#Images &Resolution (m) & Image width &GL/IT/SP  & Year \\  
			\hline
			TAS~\cite{TAS}                    & HBB     & 1     & 1,319      & 30   &--    & 792           &\ding{53}\,\ding{53}\,\ding{53}    & 2008 \\
			OIRDS~\cite{OIRDS}                & OBB     & 5     & 1,800      & 900  &up to 0.08    & 256 to 640    &\checkmark\,\checkmark\,\checkmark    & 2009 \\
			SZTAKI-INRIA~\cite{SZTAKI-INRIA}  & OBB     & 1     & 665        & 9    &--    & $\sim$800     &\ding{53}\,\ding{53}\,\ding{53}    & 2012 \\
			NWPU-VHR10~\cite{NWPUVHR}     & HBB     & 10    & 3,651      & 800      &0.08 to 2    & $\sim$1,000   &\ding{53}\,\ding{53}\,\ding{53}    & 2014 \\
			DLR-MVDA~\cite{DLR3KMunichVehicle}  & OBB     & 2     & 14,235     & 20 &0.13      & 5,616     &\ding{53}\,\ding{53}\,\checkmark        & 2015 \\
			UCAS-AOD~\cite{ucas-aod}          & OBB     & 2     & 14,596     & 1,510 &--   &$\sim$1,000    &\ding{53}\,\ding{53}\,\ding{53}   & 2015 \\
			VEDAI~\cite{VEDAI}                & OBB     & 9     & 3,640      & 1,210 &0.125   & 512;1,024     &\checkmark\,\ding{53}\,\ding{53}    & 2016 \\
			COWC~\cite{COWC}                  & CP      & 1     & 32,716     & 53  &0.15     & 2,000 to 19,000  &\checkmark\,\ding{53}\,\ding{53}   & 2016 \\
			HRSC2016~\cite{HRSC2016}          & OBB     & 26    & 2,976      & 1,061 &--   & $\sim$1,100   &\ding{53}\,\ding{53}\,\ding{53}    & 2016 \\
			RSOD~\cite{RSOD}                  & HBB     & 4     & 6,950      & 976  &0.3 to 3    & $\sim$1,000  &\ding{53}\,\ding{53}\,\ding{53}     & 2017 \\
			CARPK~\cite{CARPPK}              & HBB     & 1     & 89,777     & 1,448 &--   & 1,280     &\ding{53}\,\ding{53}\,\checkmark         & 2017 \\
 			SSDD/SSDD+~\cite{SSDD}    &HBB/OBB    &1    &2,456     &1,160   &1 to 15  &$\sim$500     &\ding{53}\,\ding{53}\,\checkmark    & 2017 \\
			SpaceNet1-6*~\cite{SpaceNet}          &Polygon  &1      & 859,982    & --    &up to 0.3   & --      &\checkmark\,\checkmark\,\checkmark          &2018 \\
			LEVIR~\cite{LEVIR}                & HBB     & 3     & 11,028     & 22,000 &0.2 to 1  & 800       &\ding{53}\,\ding{53}\,\ding{53}          & 2018 \\
			VisDrone~\cite{visdrone}          & HBB     & 10    & 54,200     & 10,209 &--  & 2,000           &\ding{53}\,\ding{53}\,\ding{53} & 2018 \\ 
			xView~\cite{xview}                & HBB     & 60    & 1,000,000  & 1,413  &0.3  & $\sim$3,000   &\checkmark\,\ding{53}\,\checkmark    & 2018 \\
			DOTA-v1.0~\cite{DOTA}             & OBB     & 15    & 188,282	 & 2,806  &up to 0.3  & 800 to 13,000  &\ding{53}\,\ding{53}\,\ding{53}      & 2018 \\ 
			ITCVD~\cite{ITCVD}                & HBB     &1      & 29,088     & 173    &0.1  & 3,744;5,616        &\ding{53}\,\ding{53}\,\ding{53} &2018 \\
			WHU building dataset~\cite{ji2018fully} &Polygon  &1  &221,107  &25,420  &0.075 to 2.7 &512 &\ding{53}\,\ding{53}\,\ding{53}  &2018\\
			DeepGlobe Building~\cite{DGB}        &Polygon  &2  &302,701  &24,586 &0.3  &650  &\ding{53}\,\ding{53}\,\checkmark &2018\\
			OpenSARShip~\cite{OpenSARShip}        &Chip  &1  &11,346  &41   &$\sim$10  & -- &\checkmark\,\checkmark\,\checkmark &2018\\
			CrowdAI Mapping Challenge~\cite{CrowdAI}    &Polygon  &1  &2,910,917  &341,058   &--  &300 &\ding{53}\,\ding{53}\,\ding{53} &2018\\
			Airbus Ship Detection Challenge~\cite{AirbusShip}    &Polygon  &1  &$\sim$131,000  &208,162   &--  &768   &\ding{53}\,\ding{53}\,\ding{53} &2018\\
			iSAID~\cite{iSAID,DOTA}     & Polygon     & 15    & 655,451	 & 2,806  &up to 0.3  & 800 to 4,000  &\ding{53}\,\ding{53}\,\ding{53}      & 2019 \\ 
			HRRSD~\cite{HRRSD2019}            & HBB     & 13    & 55,740	 &21,761  &0.15 to 1.2  & 152 to 10,569     &\ding{53}\,\ding{53}\,\ding{53}  & 2019 \\ 
			DIOR~\cite{DIOR2019}              & HBB     & 20    & 192,472    & 23,463 &0.5 to 30  & 800      &\ding{53}\,\ding{53}\,\ding{53}         & 2019 \\ 
			DOTA-v1.5~\cite{DOTA2}             & OBB     & 16    & 402,089	 & 2,806  &up to 0.3  & 800 to 13,000  &\ding{53}\,\ding{53}\,\ding{53}     & 2019 \\ 
			SAR-Ship-Dataset~\cite{SARShipDataset}      &HBB     & 1    & 5,9535 	 & 43,819  &up to 3  & 256   &\ding{53}\,\ding{53}\,\checkmark    & 2019 \\ 
			AIR-SARShip~\cite{AIR-SARShip-1.0}    &HBB  &1  &2,040  &300   &1;3  &1,000   &\checkmark\,\checkmark\,\checkmark &2020\\
			HRSID~\cite{HRSID}             & HBB     & 1    & 16,951	 & 5,604   &0.5;1;3  & 800   &\ding{53}\,\ding{53}\,\checkmark    & 2020 \\ 
			RarePlanes~\cite{RarePlanes}             & Polygon     & 1    & 644,258	 & 50,253 &0.3   &--  &\checkmark\,\ding{53}\,\checkmark     & 2020 \\
			DOTA-v2.0~\cite{DOTA2}             & OBB     & 18    & 1,793,658	 & 11,268 &up to 0.3  & 800 to 20,000   &\ding{53}\,\ding{53}\,\ding{53}    & 2020 \\  
			\hline
		\end{tabular}
		\begin{tablenotes}
    	\scriptsize
    	    \item[*] For simplicity, we summarize the~{\em SpaceNet1$\sim$6} as a whole, considering their common functionality for building detection. Note that~{\em SpaceNet3/5} are also associated with road network detection.~{\em SpaceNet7}~\cite{SpaceNet} with 11,080,000 and {\em xBD}~\cite{xBD} with 850,736 building footprints (referenced in Table~\ref{tab:CDdataset}) can also be used for building object detection and instance segmentation. {\em CrowdAI Mapping Challenge} is presented with the train and validation sets for their accessibility. {\em Annot.} refers to the {\em Annotation} style of instances, \ie, {\em HBB} ({\em Horizontal Bounding Box}) and {\em OBB} ({\em Oriented Bounding Box}). {\em CP} refers to the annotation with only the {\em Center Point} of an instance.
    	\end{tablenotes}
	  \end{threeparttable}
    \vspace{-3mm}
\end{table*}

\subsubsection{Categories Involved in Interpretation} 
The interpretation of RS images aims to extract content of interest at pixel-, region-, and scene-level. Usually, the category information of image content is extracted through elaborately designed interpretation algorithms. Hence, some datasets are constructed to recognize common RS scenes~\cite{BSC2015,SAT-6,RSSCN7,EuroSAT,RSC11,SIRI,UCM,xia2010structural} in the earlier years. To extract specific information of objects, there are datasets focusing on one or several main categories~\cite{TAS, SZTAKI-INRIA, VEDAI,DLR3KMunichVehicle,ucas-aod,COWC,HRSC2016,RSOD,CARPPK,LEVIR,RarePlanes}, such as vehicle~\cite{TAS,DLR3KMunichVehicle,COWC,CARPPK,ucas-aod}, building~\cite{SZTAKI-INRIA,xBD,ji2018fully,DGB,CrowdAI}, airplane~\cite{ucas-aod,LEVIR, RarePlanes}, and ship~\cite{SSDD,LEVIR,SARShipDataset,AirbusShip,OpenSARShip,HRSID}. The determination of semantic categories plays a significant role in real applications like land classification, urban planning, and environmental monitoring. Hence, a number of datasets are annotated for the purpose of land use and land cover (LULC) or agriculture application~\cite{UCM,LandCoverNet,LandCover.ai,Tong2019GID,Agriculture-Vision,Dstl-SIFD,Hi-UCD}. There are many semantic segmentation datasets that concern specific categories like building and road~\cite{MnihThesis, maggiori2017can, ji2018fully,shakeel2019deep,CITY-OSM}, cloud~\cite{SPARCSValidation,95-cloud,L8CloudMask,BaetensCloudValidation,S2CMC}. Some datasets aim to interpret multiple land-cover categories within specific areas, \eg, city areas~\cite{Pavia,PaviaSite,ISPRS-Vaihingen-Potsdam,ZurichSummer,Hi-UCD,S2MTCP}, relating to intensive human activities. Even with accurate annotation of category information, these datasets are with relatively small numbers of interpretation categories, which can be used for content interpretation when certain specific objects are concerned. 
 
It is obvious that the above mentioned datasets prefer to advance interpretation algorithms with limited semantic categories. However, there are more semantic categories in practical applications of RS image interpretation. As compensation for this situation, a lot of RS image datasets have been paid efforts to annotate dozens of semantic categories of interest, such as NWPU-RESISC45~\cite{RESISC45}, AID~\cite{AID}, RSI-CB~\cite{RSI-CB}, RSD46-WHU~\cite{RSD46-WHU}, Patternet~\cite{PatternNet}, Optimal-31~\cite{optimal31}, fWoM~\cite{fWoM}, CLRS~\cite{CLRS}, MLRSNet~\cite{MLRSNet}, xVew~\cite{xview}, SEN12MS~\cite{SEN12MS2019} and SECOND~\cite{SECOND}, SkyScapes~\cite{Skyscapes}, emphasizing broadly on scene-, object-, and pixel-level information. Even with enriched semantic categories, to fully interpret the content of interest in RS images still remains difficult. Take the LULC application as an example, there are a number of semantic categories enven hundreds of fine-grained classes. As a result, datasets with the limited number of scene categories are not able to extract the various and complex semantic content reflected in RS images. Moreover, categories in these datasets are set equal while the relationship between different categories, \eg, the including, included or cross relationship, is ignored. This inevitably results in the chaotic category organization and management for semantic information. Particularly, the intra-class and inter-class relationships are simply neglected in many datasets. Not only that, the context which can reveal the relationship between content of interest and their surrounding environment is rarely considered. Encouragingly, the significant exploration of relation modeling methods for RS image interpretation has been developed to address these issues~\cite{hong2020graph}. Nevertheless, how to annotate datasets with rich semantic categories and reasonable relationship organization strives to be a key problem for practical dataset construction. 

\begin{table*}[!ht]
\centering
\caption{Comparison of different RS image semantic segmentation datasets.}\label{tab:SSdataset}
\vspace{-1mm}
\begin{threeparttable}
\setlength{\tabcolsep}{1.2mm}
\begin{tabular}{lcccccccc}
	\hline
	Datasets                  &\#Cat.  &\#Images    &Resolution (m)    &\#Channels    &Image size &GL/IT/SP   &Year\\
	\hline
	Kennedy Space Center \cite{KSC} &13    &1    &18    &224   &512$\times$614  &\ding{53}\,\checkmark\,\checkmark  &2005\\
	Botswana \cite{KSC} &14    &1    &30    &242    &1,476$\times$256  &\ding{53}\,\checkmark\,\checkmark  &2005\\
	Salinas \cite{PaviaSite}  &16    &1    &3.7   &224    &512$\times$217  &\ding{53}\,\ding{53}\,\checkmark  &--\\
	University of Pavia  \cite{PaviaSite} &9  &1  &1.3  &115  &610$\times$340  &\ding{53}\,\ding{53}\,\checkmark  &--\\
	Pavia Centre \cite{PaviaSite}&9  &1  &1.3  &115 bands  &1,096$\times$492  &\ding{53}\,\ding{53}\,\checkmark   &--\\
	ISPRS Vaihingen \cite{ISPRS-Vaihingen-Potsdam}&6 &33  &0.09  &IR,R,G,DSM,nDSM  &$\sim$2,500$\times$2,500 &\ding{53}\,\ding{53}\,\checkmark &2012\\
	ISPRS Potsdam \cite{ISPRS-Vaihingen-Potsdam} &6  &38  &0.05  &IR,RGB,DSM,nDSM  &6,000$\times$6,000  &\checkmark\,\ding{53}\,\checkmark  &2012\\
	Massachusetts Buildings \cite{MnihThesis}  &2    &151       &1      &RGB    &1,500$\times$1,500 &\checkmark\,\checkmark\,\ding{53} &2013\\
	Massachusetts Roads \cite{MnihThesis}      &2    &1,171    &1      &RGB    &1,500$\times$1,500 &\checkmark\,\checkmark\,\ding{53}  &2013\\
	Indian Pines \cite{IndianPines}          &16    &1    &20    &224    &145$\times$145    &\checkmark\,\checkmark\,\checkmark   &2015\\
	Zurich Summer \cite{ZurichSummer}    &8    &20    &0.62    &NIR, RGB     &1,000$\times$1,150 &\checkmark\,\checkmark\,\checkmark   &2015\\
    SPARCS Validation~\cite{SPARCSValidation} &7    &80    &30    &11     &1,000$\times$1,000  &\checkmark\,\checkmark\,\checkmark  &2016\\
    Biome~\cite{L8CloudMask} &4    &96    &30    &11     &$\sim$9,000$\times$9,000 &\checkmark\,\checkmark\,\checkmark    &2017\\
	Inria~\cite{maggiori2017can}    &2    &360       &0.3    &RGB    &5,000$\times$5,000  &\ding{53}\,\ding{53}\,\ding{53}  &2017\\
    EvLab-SS \cite{EVlab-SS}  &10    &60       &0.1 to 2    &RGB    &4,500$\times$4,500 &\ding{53}\,\ding{53}\,\checkmark   &2017\\
    RIT-18 \cite{RIT-18}  &18    &3       &0.047    &6    &9,000$\times$6,000  &\checkmark\,\checkmark\,\checkmark  &2017\\
	CITY-OSM~\cite{CITY-OSM}  &3    &1,671       &~0.1    &RGB   &2,500$\times$2,500 to 3,300$\times$3,300  &\ding{53}\,\ding{53}\,\ding{53}  &2017\\
	Dstl-SIFD*\cite{Dstl-SIFD}  &10    &57       &up to 0.3   &up to 16   &$\sim$3,350$\times$3,400  &\checkmark\,\ding{53}\,\checkmark  &2017\\
	IEEE GRSS Data Fusion Contest 2017  &17    &30       &1,4    &9    &643$\times$666;374$\times$515 &\checkmark\,\checkmark\,\checkmark   &2017\\
	IEEE GRSS Data Fusion Contest 2018  &20    &1    &1    &48    &4,172$\times$1,202  &\checkmark\,\checkmark\,\checkmark  &2018\\
	Aeroscapes~\cite{aeroscapes} &11    &3,269    & --    &RGB    &720$\times$1,280 &\ding{53}\,\ding{53}\,\ding{53}   &2018\\
	DLRSD~\cite{DLRSD}  &17  &2,100  &0.3  &RGB  &256$\times$256 &\ding{53}\,\ding{53}\,\ding{53} &2018\\
	DeepGlobe Land Cover~\cite{DGB}  &7  &1,146  &0.5  &RGB  &2,448$\times$2,448 &\ding{53}\,\ding{53}\,\checkmark &2018\\	
	So2Sat LCZ42~\cite{So2SatLCZ42}  &17    &400,673       &10    &10    &32$\times$32  &\checkmark\,\ding{53}\,\checkmark  &2019\\
	SEN12MS~\cite{SEN12MS2019}  &33    &180,662 triplets       &10 to 50    &up to 13    &256$\times$256  &\checkmark\,\ding{53}\,\checkmark  &2019\\
	95-Cloud~\cite{95-cloud} &1    &43,902     &30    &NIR,RGB    &384$\times$384  &\checkmark\,\ding{53}\,\checkmark  &2019\\
	Shakeel et al.~\cite{shakeel2019deep} &1  &2,682  &0.3  &RGB  &300$\times$300 &\ding{53}\,\ding{53}\,\ding{53} &2019\\
	ALCD Cloud Masks~\cite{BaetensCloudValidation} &8    &38     &10    &RGB    &1,830$\times$1,830 &\checkmark\,\checkmark\,\checkmark   &2019\\
	SkyScapes~\cite{Skyscapes} &31    &16     &0.13    &RGB    &5,616$\times$3,744  &\ding{53}\,\ding{53}\,\ding{53}  &2019\\
	DroneDeploy~\cite{DroneDeploy} &7    &55     &0.1    &RGB    &up to 12,039$\times$13,854  &\ding{53}\,\ding{53}\,\ding{53}  &2019\\
	Slovenia LULC~\cite{Slovenia} &10    &940     &10    &6    &5,000$\times$5,000  &\checkmark\,\checkmark\,\checkmark  &2019\\
	LandCoverNet~\cite{LandCoverNet}  &7  &1,980  &10 &NIR,RGB	    & 256$\times$256 &\checkmark\,\checkmark\,\checkmark & 2020 \\ 
	UAVid \cite{lyu2018uavid}  &8    &420       &--    &RGB    &$\sim$4,000$\times$2,160  &\ding{53}\,\ding{53}\,\checkmark  &2020\\
	GID \cite{Tong2019GID}  &15    &150       &0.8 to 10    &4    &6,800$\times$7,200  &\checkmark\,\checkmark\,\checkmark  &2020\\
	LandCover.ai~\cite{LandCover.ai} &3  &41  &0.25,0.5  &RGB  &9,000$\times$9,500;4,200$\times$4,700 &\checkmark\,\ding{53}\,\ding{53} &2020\\
	Agriculture-Vision~\cite{Agriculture-Vision}  &9  &94,986  &0.1;0.15;0.2  &NIR,RGB  &512$\times$512  &\ding{53}\,\ding{53}\,\checkmark & 2020 \\
	S2CMC*~\cite{S2CMC}  &18  &513  &20  &13  & 1,024$\times$1,024  &\checkmark\,\checkmark\,\checkmark &2020 \\
	\hline
\end{tabular}
\begin{tablenotes}
	\scriptsize
	    \item[*] The UAVid consists of 30 video sequences captured by unmanned aerial vehicle and each sequence is annotated by every 10 frames, resulting in 420 densely annotated images. The~{\em S2CMC} is short for~{\em Sentinel-2 Cloud Mask Catalogue}. The~{\em DSTL-SIFD} is short for the challenge of~{\em Dstl Satellite Imagery Feature Detection}.
\end{tablenotes}
 \end{threeparttable}
\end{table*}

\begin{table*}[!t]
\centering
\caption{Comparison of different RS Image change detection datasets.}\label{tab:CDdataset}
\vspace{-1mm}
  \setlength{\tabcolsep}{1.7mm}
  \begin{tabular}{lcccccccc}
	\hline
	Datasets  &\#Cat.    &\#Image pairs  &Resolution (m)  &\#Channels  &Image size &GL/IT/SP   &Year\\
	\hline
	SZTAKI AirChange~\cite{benedek2009change}  &2  &13  &1.5    &RGB  &952$\times$640 &\ding{53}\,\checkmark\,\ding{53} &2009\\
	AICD~\cite{bourdis2011constrained}&2  &1,000 &0.5 &115 &800$\times$600 &\ding{53}\,\ding{53}\,\ding{53} &2011\\
	Taizhou Data~\cite{wu2013slow}  &4  &1  &30  &6 &400$\times$400  &\checkmark\,\checkmark\,\checkmark &2014\\
	Kunshan Data~\cite{wu2013slow}  &3  &1  &30  &6 &800$\times$800 &\checkmark\,\checkmark\,\checkmark &2014\\
	Cross-sensor Bastrop~\cite{BastropFireDataeset}  &2  &4  &30,120  &7,9 &444$\times$300; 1,534$\times$808 &\checkmark\,\checkmark\,\checkmark  &2015\\
	MtS-WH~\cite{MtS-WH2017}  &9  &1  &1  &NIR, RGB  &7,200$\times$6,000 &\checkmark\,\checkmark\,\checkmark &2017\\
	Yancheng~\cite{song2018change}  &4  &2  &30  &242 &400$\times$145  &\checkmark\,\checkmark\,\checkmark &2018\\
	GETNET dataset~\cite{8418840} &2  &1  &30  &198 &463$\times$241  &\ding{53}\,\checkmark\,\checkmark &2018\\
	Urban-rural boundary of Wuhan~\cite{he2018land}  &20  &1  &4/30  &4, 9  &960$\times$960 &\checkmark\,\checkmark\,\checkmark &2018\\
	Hermiston City, Oregon~\cite{lopez2018stacked}  &5  &1  &30  &242 &390$\times$200  &\checkmark\,\checkmark\,\checkmark & 2018\\
	OSCD~\cite{daudt2018urban}   &2  &24  &10  &13  &600$\times$600  &\checkmark\,\checkmark\,\checkmark &2018\\
	WHU building dataset \cite{ji2018fully}  &2  &1  &0.2  &RGB  &32,507$\times$15,354 &\checkmark\,\checkmark\,\checkmark &2018\\
	Season-varing dataset~\cite{SeasonVaring2018}  &2  &16,000  &0.03 to 0.1  &RGB  &256$\times$256 &\ding{53}\,\ding{53}\,\ding{53} &2018\\
    ABCD~\cite{ABCD2017}  &2  &16,950  &0.4  &RGB  &128$\times$128;160$\times$160  &\ding{53}\,\checkmark\,\ding{53} &2018\\
	California flood dataset~\cite{CaliforniaFloodDataset}  &2  &1  &5,30  &RGB,11  &1534$\times$808 &\checkmark\,\checkmark\,\checkmark &2019\\
	L{\'o}pez-Fandi{\~n}o et al.~\cite{HCDD}  &5  &2  &20  &224 &984$\times$740; 600$\times$500 &\ding{53}\,\checkmark\,\checkmark &2019\\
	xBD~\cite{xBD}  &6  &11,034  &~up to 0.8  &RGB  &1,024$\times$1,024  &\checkmark\,\checkmark\,\checkmark &2019\\
	HRSCD~\cite{HRSCD2019}  &6  &291  &0.5  &RGB  &10,000$\times$10,000  &\checkmark\,\checkmark\,\checkmark &2019\\
	LEVIR-CD~\cite{LEVIR-CD2020}  &2  &637  &0.5  &RGB  &1,024$\times$1,024  &\ding{53}\,\ding{53}\,\ding{53} &2020\\
	SECOND~\cite{SECOND}  &30  &4,214  &0.5 to 3  &RGB  &512$\times$512  &\ding{53}\,\ding{53}\,\ding{53} &2020\\
	Google Dataset~\cite{GoogleDataset}  &2  &1,067  &0.55  &RGB  &256$\times$256 &\checkmark\,\checkmark\,\ding{53} &2020\\
	Zhang et al.~\cite{9052762} &2  &4  &2;2.4;5.8  &NIR, RGB  &1,431$\times$1,431; 458$\times$559; 1,154$\times$740 &\checkmark\,\checkmark\,\checkmark  &2020\\
	Hi-UCD~\cite{Hi-UCD} &9  &1,293  &0.1  &RGB  &1,024$\times$1,024 &--/--/Y  &2020\\
	SpaceNet7\cite{SpaceNet} &--  &24  &4  &RGB  &-- &\checkmark\,\checkmark\,\checkmark  &2020\\
	S2MTCP~\cite{S2MTCP} &2  &1,520  &up to 10  &13  &600$\times$600  &\checkmark\,\checkmark\,\checkmark  &2021\\
	\hline
  \end{tabular}
\end{table*}

\subsubsection{Dataset Annotation} 
To our knowledge, most of the datasets listed in Tables~\ref{tab:SCdataset}-\ref{tab:CDdataset} are manually annotated by experts. Generally, the work of dataset annotation is to assign semantic tags to scenes, objects or pixels of interest in RS images. For the task of scene classification, a category label is typically assigned to the scene components by visual interpretation of experts~\cite{AID,RESISC45}. In order to recognize specific objects, entities in images are usually labeled with closed areas. Thus, many existing datasets manually annotate objects in the form of bounding boxes,~{\eg, NWPU-VHR10~\cite{NWPUVHR}, RSOD~\cite{RSOD}, HRRSD~\cite{HRRSD2019}, and DIOR~\cite{DIOR2019}}, or enclosed polygons,~\eg,~iSAID~\cite{iSAID} and xBD~\cite{xBD}. Before annotating content of interest, a fundamental issue is the acquisition of target RS images in which the intriguing content is contained. Usually, the target images are manually searched, distinguished, and screened in the image database by trained annotators. Along with the subsequent label assignment, the whole annotation process in the construction of RS image datasets is time-consuming and labor-intensive, especially for the pixel-wise annotations as shown in Tables~\ref{tab:SSdataset}-\ref{tab:CDdataset}. As a result, dataset construction, from source image collection, semantic information annotation, and quality review, relies heavily on manual operations, making it an expensive project. This raises an urgent demand for developing more efficient and assistant strategies to lighten the burden of artificial annotation.

When it comes to the annotation tools, there is a lack of visualization methods for the annotation of large scale and hyper-spectral RS images. Currently, annotation tools designed for natural images, \eg, LabelMe~\cite{LabelMe} and LabelImg~\cite{LabelImg}, are introduced to annotate RS images. These annotation tools typically visualize an image with a limited scale. However, different from natural images, RS images taken from the bird-view are with large scale and wide geographic coverage. Thus, the annotator can only conduct the labeling operations within a local region of the RS image. In this situation, inaccurate annotation could be produced since it is difficult for the annotator to grasp the global content of the RS image. Meanwhile, the image roam process will inevitably constrain annotation efficiency. This problem is particularly serious when conducting annotation for semantic segmentation and change detection tasks where labels are typically assigned pixel-by-pixel~\cite{ISPRS-Vaihingen-Potsdam,IndianPines,benedek2009change,lyu2018uavid}. On the other hand, hyper-spectral RS images~\cite{KSC,Pavia,ISPRS-Vaihingen-Potsdam,IndianPines,song2018change,he2018land,lopez2018stacked,daudt2018urban} which characterize objects with rich spectral signatures, are usually employed for elaborate interpretation of semantic content. However, it is hard to label the hyper-spectral RS images since annotation tools developed for natural images are not able to visualize hyper-spectral images of hundreds of spectral bands. Therefore, universal annotation tools are desperately desired to be developed for efficient and convenient semantic annotation, especially for the large scale and hyper-spectral RS images.

\subsubsection{Image Source} 
A wide group of RS images has been employed as the source of interpretation datasets, including the optical, multi-/hyper-spectral, SAR images. Typically, the optical images from Google Earth are widely employed as the data standard, such as those for scene classification~\cite{AID,UCM,RSSCN7,RSC11,SIRI,RESISC45,RSD46-WHU,PatternNet}, object detection~\cite{ding2019learning,TAS,ucas-aod,HRSC2016,RSOD,LEVIR,DOTA,DIOR2019}, and pixel-level analysis~\cite{MnihThesis,ZurichSummer,maggiori2017can}. In these scenarios, RS images are typically interpreted by the visual content, of which the spatial pattern, texture structure, information distribution as well as organization mode are more concerned. Although the Google Earth images are post-processed with RGB formats using the original optical aerial images, they possess the potential for pixel-based LULC interpretation as there is no general statistical difference between the Google Earth images and optical aerial images~\cite{rs5116026}. Thus, Google Earth images can also be used as RS images for evaluating interpretation algorithms~\cite{AID}.

Different from the optical RS image datasets, the construction of hyper-spectral and SAR image datasets should adopt the original data formats. Compared to optical images, multi-/hyper-spectral images can capture the essential characteristics of ground features as the rich spectral and spatial information are simultaneously involved. Therefore, the content interpretation of hyper-spectral RS images is mainly based on the spectral properties of ground features. Naturally, this kind of images is typically employed to construct the dataset for subtle semantic information extraction, such as semantic segmentation~\cite{KSC,Pavia,ISPRS-Vaihingen-Potsdam,IndianPines,So2SatLCZ42,SEN12MS2019} and change detection~\cite{zhang2018coarse,song2018change,he2018land,lopez2018stacked,daudt2018urban,S2MTCP}, where more attention is paid to the knowledge of the fine-grained compositions. For SAR images acquired by microwave imaging, content interpretation is usually performed by the radiation, transmission, and scattering properties. Hence, SAR images are employed for abnormal object detection by utilizing the physical properties of ground features. And it is not encouraged to employ the modified data of SAR images for visual interpretation of interested content. It is worth noting that the advantages of different RS images can be integrated. This is why the multi-modal learning framework has drawn much attention and been employed to greatly improve the performance of RS image interpretation~\cite{hong2020more}, which provide significant reference for making the most of different RS image datasets, especially those from different imaging sensors.

\subsubsection{Dataset Scale}  
A large number of RS image datasets have been constructed for various interpretation tasks. However, many of them are with small scales, reflected in aspects like the limited number, small size, and lacked diversity of annotated images. On the one hand, the size and number of images are important properties concerning the scale of a RS image dataset. RS images that typically taken from the bird-view perspective have a large geographic coverage and thus possess large image size. For example, an image from GF-2 satellite usually exceeds $30,000\times30,000$ pixels. However, many of the current datasets employ the chipped images, usually with the width/height of a few hundred pixels as shown in Tables~\ref{tab:SCdataset}-\ref{tab:CDdataset}, to fit specific models that are designed to extract features within the limited scale of images. In fact, the preservation of the original image size is more close to real-world applications~\cite{DOTA,Tong2019GID}. Some datasets with larger image sizes, say, width/height of a few thousand pixels, are limited with the number of annotated images or categories~\cite{COWC,RIT-18,HRSCD2019,MtS-WH2017,SeasonVaring2018,LEVIR-CD2020}. Furthermore, quite a few datasets contain one or several images, especially those for semantic segmentation~\cite{KSC,Pavia,IndianPines} and change detection~\cite{wu2013slow,BastropFireDataeset,song2018change,8418840,he2018land,lopez2018stacked,ji2018fully,CaliforniaFloodDataset,HCDD,MtS-WH2017,9052762}, which are limited by the high cost of pixel-wise annotation. As a result, the scale limitations in size and number of images could easily lead to performance saturation for interpretation algorithms. 

On the other hand, due to the constraint of data scale, existing datasets often show deficiencies in image variation and sample diversity. Typically, content in RS images always shows differences with the change of spatio-temporal attributes while images in some of the datasets are selected from local areas or with limited imaging conditions~\cite{BSC2015,RSD46-WHU,DLR3KMunichVehicle,KSC}. In addition, content reflected in RS images are with complex texture, structure, and spectral features owing to the high complexity of the Earth's surface. Thus, datasets with limited images and samples~\cite{UCM,RSC11,SZTAKI-INRIA,COWC,KSC,RIT-18} are usually not able to completely characterize the properties of objects of interest. As a result, there is a lack of representativeness of real-world scenarios for datasets with small scales. This can lead to weak interpretation ability of algorithms with the change of application scenarios. Furthermore, constrained by the scale of datasets, the currently popular deep learning approaches are usually pre-trained using the large-scale natural image datasets, \eg, ImageNet~\cite{Imagenet}, and then used for RS image interpretation~\cite{SCdeep6,SCdeep5}. Nevertheless, features learned by this strategy are hard to completely adapt to RS data because of the essential difference between RS images and natural images. For instance, the change of object orientation is common to be observed in RS images. All of these raise an urgent demand for annotating large-scale RS datasets with rich images to advance RS image interpretation.

% \section{How to Build a Useful Remote Sensing Image Dataset}
\section{Guidances of Building RS Image Benchmarks} \label{Conception}
The availability of a good RS image dataset has been shown critical for effective feature learning, algorithm development, and high-level semantic understanding~\cite{Imagenet,PASCALVOC,COCO,cordts2016cityscapes,Zhu2016}. More than that, the performance of almost all data-driven methods rely heavily on the training dataset. However, constructing a large-scale and meaningful image dataset for RS image interpretation is not an easy job, at least from the points of technology and cost factors. The challenge lies largely in the aspect of efficiency and quality control. The absence of systematic work involving these problems has limited the construction of practical datasets and continuous advancement of interpretation algorithms in RS community. Therefore, it is valuable to explore the feasible scheme for creating a practical RS image dataset. We believe that the following introduced aspects can be taken into account when creating a desirable dataset for RS image interpretation.

   \subsection{Desirable Properties of Benchmark Datasets}
     In order to enhance the practicality, the dataset for RS image interpretation should be created toward practical application requirements rather than the characteristics of interpretation algorithms. Essentially, the creation of RS image dataset aims at model training, testing, and screening for practical applications. It is of great significance to get the whole picture of a designed interpretation model before it is poured into practical applications. Thus, the reliable benchmark dataset becomes critical to comprehensively verify the validity of designed interpretation model. To this end, the created dataset should consist of sufficient and accurately annotated samples that cover the challenges in practical application scenarios. 
     
     In this point of view, the annotation of RS image dataset is better to be conducted by the application sides rather than the algorithm developers. Annotations by algorithm developers will inevitably possess bias as they may be more familiar with the algorithm properties and lack of understanding of challenges lying in practical applications. As a result, the annotated dataset from developers could be at risk of being algorithm-oriented. On the contrary, the application sides have more opportunities to access the real application scenarios, and thus, are more familiar with the issues and challenges lying in the interpretation tasks. Therefore, the dataset annotation from application sides is more reliable, and thus, conducive to enhance the practicability of the interpretation algorithm.

\begin{figure*}[htb!]
      \centering
      \includegraphics[width=0.8\linewidth]
      {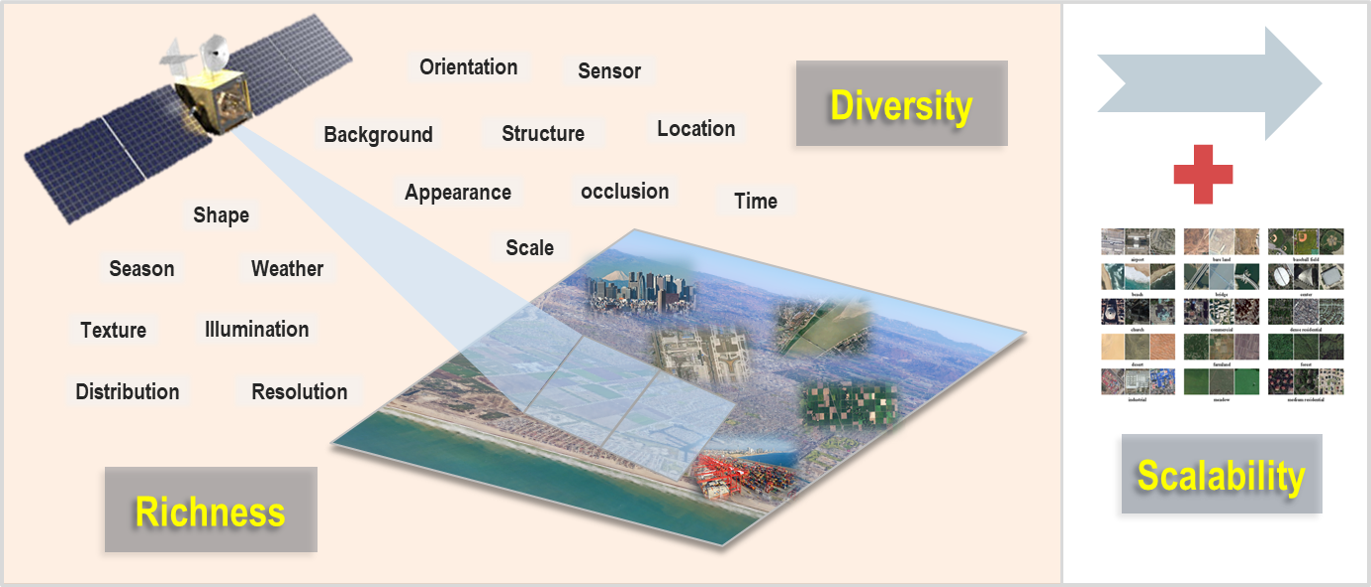}
      \caption{The {\em \textbf{DiRS}} properties: {\em \textbf{di}versity}, {\em \textbf{r}ichness}, and {\em \textbf{s}calability}. DiRS formulates the basic properties which can be considered as basic desirability in the construction of datasets for RS image interpretation. We believe that these properties are complementary to each other. That is, the improvement of dataset in one aspect can simultaneously promote the dataset quality reflected in other properties.} 
      \label{figure:DiRS}
      \vspace{-3mm}
    \end{figure*}

     In general, the RS image dataset should be constructed toward the real-world scenarios instead of the specific algorithms. Thus, it is possible to feed the interpretation system with high-quality data, which boost the interpretation algorithms to effectively learn and even extend knowledge that people desired. With these points in mind, we believe that the {\em \textbf{di}versity}, {\em \textbf{r}ichness}, and {\em \textbf{s}calability} (called {\em \textbf{DiRS}}), as illustrated in Figure~\ref{figure:DiRS}, could be considered as the desirable properties when creating benchmark datasets for RS image interpretation. 
   
    \subsubsection{Diversity}
      A dataset is considered to be diverse if its annotated objects depict various visual characteristics of relevant semantic content with a certain degree of complementarity. From the perspective of within-class diversity, annotated objects with large diversity are able to comprehensively represent content distribution in real world. To this end, it is better that each annotated object could reflect different attributes rather than the repeated characteristics. For example, the annotated objects in the same category, \eg, vehicle, can be distinguished from each other in properties like appearance, scale, and orientation that diversify the instances. Thus, the within-class objects of large diversity are conducive for an algorithm to learn the essential characteristics. In addition, it should be emphasized that the imaging and geographic properties is also desperately desired for dataset diversity improvement. In the real world, the properties of objects of interest could vary with its geographic location and imaging time. A fact is that the object of the same class could show differences in state, surroundings, and position with the spatio-temporal property variation. Thus, the imaging and geographic properties become nonnegligible when building an interpretation dataset of high diversity. Especially, this is very important for the large-scale geographic application using method learned from a given dataset. Therefore, annotated objects with these distinct characteristics are able to provide insurance of training interpretation model with powerful ability of feature representation and application generalization. In this regard, the within-class diversity actually emphasizes on the individual differences between objects of interest in the same class. 
      
      On the other hand, in order to learn an interpretation algorithm for effective discrimination of different classes, the between-class diversity should also be taken into consideration when constructing the RS image dataset. For this requirement, the fine-grained classes, particularly those with high semantic overlapping, should be contained as many as possible. Objects of different semantic classes usually take specific feature pattern and distribution. Thus, annotating objects with diverse semantic classes can enable an interpretation model to learn more powerful feature representation. Besides, high semantic overlapping in different categorical objects means large between-class similarity. It is easy to understand that the notable intervals of content features can make an interpretation model learn to distinguish different classes effortlessly. In contrast, objects with high semantic overlapping, denoting the small distance of different classes, will put forward higher requirements for interpretation models to discriminate similar semantic content. From this point of view, the between-class diversity pays more attention to the common characters among objects of different classes. Generally, the within-class and between-class diversity simultaneously offer the guarantee for feature complementarity and peculiarity for annotated objects, which is crucial for constructing datasets of large diversity.

    \subsubsection{Richness}
    In addition to the diversity that emphasises on the otherness of objects, the richness of a dataset is another significant property, which attaches importance to the variation of images. Specifically, the rich image variation regards various content characteristics and large-scale samples as important when constructing a RS image dataset.
    In order to enrich the content characteristics, images can be collected under various circumstances, such as the weather, season, illumination, imaging condition, and sensor, which allow the dataset to possess rich variations in translation, viewpoint, object pose, spatial resolution, illumination, background, occlusion, {\em etc}. Not only that, images collected from different periods and geographic regions can also endow the dataset with rich spatio-temporal distribution.

    Moreover, different from natural images that are usually taken from horizontal perspective with narrow extent, RS images are taken with bird-views, endowing the images with large geographic coverage, abundant ground features, and complex background information. Thus, an interpretation dataset is desired to contain images that reflect the rich characteristics,~\eg, variation in geometrical shape, structure characteristic, textural attribute, {\em etc}. From this point of view, the constructed dataset should consist of large-scale images to contain sufficient annotated samples, which is able to further ensure its comprehensive representativeness for real-world scenarios. The reality is that insufficient images and samples are more likely to lead to the over-fitting problem in model training, particularly for data-driven interpretation methods (\eg, CNN). In this regard, the scale of a RS image dataset should be large enough to ensure the richness property. Thus, the interpretation models built upon the dataset in accordance with the above lines are able to possess more powerful representation and generalization ability for practical applications.

    \subsubsection{Scalability }
    Scalability can be a measure of the ability to extend a constructed dataset. With the increasingly wide applications of RS images, the requirements for a dataset usually change along with the specific application scenarios. For example, a new category of scene may need to be differentiated from the collected categories with the change of LULC. Thus, the constructed dataset must be organized with sufficient category space to involve the new category scenes while keeping the existing category system extensible. Not only that, but the relationship among the annotated features is also better to be well managed according to the real-world application requirements. That is, a constructed benchmark dataset for RS image interpretation is better to be flexible and extendable, considering the change of application scenarios. 
    
    Notably, there is a large number of RS images received every day, which need to be efficiently labeled with valuable information to maximize their application value. To this end, the organization, preservation, and maintenance of annotations and images are of great significance to be controlled for the scalability of a dataset. Besides, it would be preferable if the newly annotated images could be involved in the constructed dataset effortlessly. Thus, the full operations of adding, updating, removing, and retrieving data and information in the constructed dataset become a significant property for scalibility. With these considerations, a constructed RS image dataset with excellent scalability can be conveniently adapted to the changing requirements for real-world applications without impacting its inherent accessibility, and thereby assuring sustainable utilization even as modifications are made.

 \subsection{Semantic Coordinates to Facilitate Image Acquisition}

   The acquisition of RS images that contain content of interest formulates the foundation of creating an interpretation dataset. Benefiting from the spatial property possessed by RS images, the RS images in the database can be accessed by utilizing their inherent information of geographic coordinates~\cite{fWoM,GeoBoost}. And further, a geographic feature is commonly presented with a series of geographic coordinates. Meanwhile, the feature is usually attached with specific tag attributes that present its semantic meaning. From this perspective, the geographic coordinates related to a specific feature element can be regarded as the semantic coordinates, by referencing the feature's tag attributes. Thus, we are able to collect the geographic coordinates and then access the corresponding tag attributes to efficiently identify the locations of RS images that contain content of interest.

   Typically, this strategy can be performed to prepare a public optical RS image dataset, by utilizing the public map application interface, open source data, and public geodatabases. The coordinates collection may not be an optimal strategy but can also be employed as a reference when creating a private dataset of which images are from other sensors and databases.

   \subsubsection{Map Search Engines}
    A convenient way to collect RS images with content of interest is to utilize public map search engines, such as Google Map\footnote{https://ditu.google.com}, Bing Map\footnote{https://cn.bing.com/maps}, and World Map\footnote{http://map.tianditu.gov.cn}. As common digital map service solutions, they provide satellite images covering the whole world in different spatial resolutions. Many existing RS datasets, such as UCM~\cite{UCM} and NWPU-RESISC45~\cite{RESISC45} for scene classification, LEVIR~\cite{LEVIR} and DOTA~\cite{DOTA} for object detection, Google Dataset~\cite{GoogleDataset} and LEVIR-CD~\cite{LEVIR-CD2020} for change detection, have been built based on Google Map. When collecting RS images on such map search engines, the developed map application programming interface (API) can be utilized to extract images and acquire the corresponding semantic tags. Based on the rich positional data composed of millions of point, line and region vectors that contain specific semantic information, the large amount of candidate RS images can be collected through these map engines. For example, by searching ``airport'' on Google Earth, all searched airports in a specific area will be indicated with specific geographic locations. The corresponding satellite images can be accessed using the coordinates of search results. Then, the acquired satellite images can be used to annotate airport scene and aircraft object samples. 

    \subsubsection{Open Source Data}
    Open source geographic data is established on the global positioning system (GPS) information, aerial photography images, other free content and even local knowledge (such as social media data) from users. Open source geographic data, such as the Open Street Map (OSM) and WikiMapia, are created upon the collaboration plan which allows users to label and edit the ground feature information. Therefore, the open source geographic data can provide rich semantic information that is timely updated, low cost and has a large amount in quantity compared with the manual collection strategy for RS images~\cite{CITY-OSM,RSI-CB}. With the abundant geographic information provided by various open source data, we are able to collect elements of interest like points, lines, and regions with specific geographic coordinates. Then, we can match the collected geographic elements with their corresponding RS images. Moreover, the extracted geographic elements of interest can be aligned with temporal RS images which can be downloaded from different map engines as described above. With these advantages and operations, it is possible to collect large-scale RS images of great diversity for dataset construction.

    \subsubsection{Geodatabase Integration}
    Different from the collection of natural images, which can be conveniently accessed through web crawling, search engines (\eg, Google image search), and sharing databases (\eg, Instagram, Flickr), the acquisition of RS images that contain content of interest is difficult because of the high searching cost. Nevertheless, the public geodatabases and geographic information products released by state institutions and communities usually provide accurate and rich geographic data. With this facility, the geographic coordinates attached with specific semantic information can be obtained through these databases. For example, the National Bridge Inventory (NBI)~\footnote{https://www.fhwa.dot.gov/bridge/nbi.cfm} presents detailed information of the bridges, including the geographic locations, length, material, and so on. Benefiting from this advantage, we can extract a large number of geographic coordinates of bridges for the collection of bridge images. By integrating these kinds of public geodatabases, we are able to obtain the geographic locations of RS images with specific semantic information, and thus, efficiently collect a large number of RS images that contain content of interest at relatively low cost.

  \subsection{Annotation Methodology}
    With the collected images for a specific interpretation task, annotation is performed to assign specific semantic labels to the content of interest in the images. Next, the common image annotation strategies will be introduced.

 \subsubsection{Annotation Strategies} Depending on whether human intervention is involved, the solutions to RS image annotation can be classified into three types: manual, automatic, and interactive annotation.  
      \begin{itemize}
      	\item {\noindent \bf Manual Annotation}
        The common way to create an image dataset is to employ the manual annotation strategy. The great advantage of manual annotation is its high accuracy because of the fully supervised annotation process. Based on this consideration, many RS image datasets have been manually annotated for various interpretation tasks, such as those for scene classification~\cite{UCM,AID,SIRI,RSC11}, object detection~\cite{DOTA,SZTAKI-INRIA,VEDAI} and semantic segmentation~\cite{Tong2019GID,lyu2018uavid}. Regardless of the source from which the natural or RS images are acquired, the way to annotate content in RS images is similar. And many tools have been built to relieve the monotonous annotation work. Hence, image annotation tools developed for natural images can be further introduced for RS images (typically the optical RS images) to pave the way for cost-effective construction of large-scale datasets. The resource concerning to image annotation tools will be introduced in Section~\ref{Resources}.
        
        In practice, constructing a large-scale image dataset by manual scheme is laborious and time-consuming as introduced before. For example, a number of people spent several years to construct the ImageNet~\cite{Imagenet}. To relieve this problem, crowd-sourcing annotation becomes an alternative solution that can be employed to create a large-scale image dataset~\cite{xview,fWoM,COCO} while paying efforts to its challenge with quality control. Besides, benefiting from excellent ability of image interpretation algorithms, annotators can also resort to machine learning schemes~\cite{PolygonRNN++,andriluka2018fluid}, which can be integrated as the preliminary annotation, to speed up the efficiency of manual annotation. 
    
       \item {\noindent \bf Automatic Annotation}
        In contrast to natural images, RS images are often characterized with complex structures and textures because of the spectral and spatial variation. It is difficult to annotate semantic content for annotators without domain knowledge. As a result, the manually annotated dataset is prone to have bias problem because of annotators' difference in domain knowledge, educational background, labelling skill, life experience, etc. In this situation, automatic image annotation methods are naturally employed to alleviate annotation difficulties and further reduce the cost of manual annotation~\cite{8658155}. 
        
        Automatic annotation methods reduce the cost of annotation by leveraging learning schemes~\cite{Mutilablel2020,afridi2018automated,maihami2018automatic,tian2017automatic,rs11020145,cheng2018survey}. In this strategy, a certain number of images are initialized to train an interpretation model, including the fully supervised~\cite{6330983} and weakly supervised methods~\cite{Yang2012sar,SARSWSL,blockAnnotation}. The candidate images are then poured into the established model for content interpretation and the interpretation results finally serve as annotation information. And iterative and incremental learning~\cite{li2010optimol} can be employed to filter noisy annotation and enhance the generalization ability of annotation model~\cite{HAN201823,dang2019open,tasar2019incremental,SARSWSL}. Nevertheless, one disadvantage of automatic annotation is that the generalization ability of the annotation model can be affected by both the quality of the initial candidate images. In addition, to decompose the difficulty of annotation and enhance the connectivity between annotation and real applications, the existing semantic information,~\eg, thematic products as a unique presentation for RS image content, can serve as the source for automatic RS image annotation and content update ~\cite{8635553}. With the inherent semantic information contained in thematic products, reliable training samples are able to be extracted~\cite{9121728}. And this idea has also been successfully employed in dataset construction,~\eg, BigEarth~\cite{BigEarthNet}, which shows promising prospect in the automatic annotation of large-scale dataset for RS image interpretation.
        
    \begin{figure*}[t!]
      \centering
      \setlength{\abovecaptionskip}{-0.01cm}
      \captionsetup{justification=centering}
      \includegraphics[width=0.85\linewidth]
          {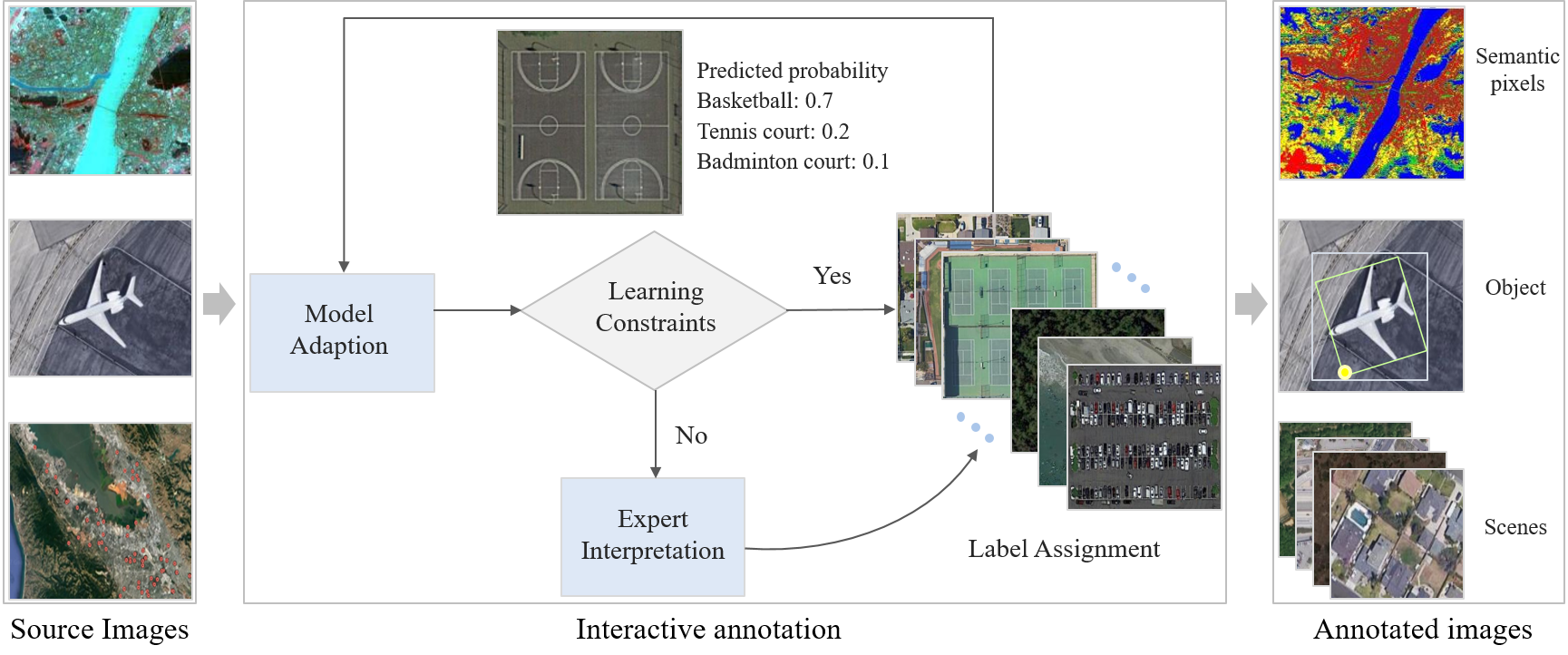}
      \caption{General workflow of Semi-automatic annotation in RS images.}
      \label{figure:semi_annotation}
      \vspace{-3mm}
    \end{figure*}

        \item {\bf Interactive Annotation}
         In the era of big RS data, annotation with human-computer interaction, which falls in semi-automatic annotation, could be a more practical strategy considering the demand for RS image annotation with high quality and efficiency. In this strategy, an initial framework can be constructed using the existing archives with available annotation and then employed to annotate the unlabeled RS images. On this basis, the performance of an annotation model can be improved greatly with the intervention from annotators~\cite{agustsson2019interactive}. The intervention from annotators can be in the form of relevance feedback or identification of the relevant content in the images to be annotated. In this scheme, the overall performance of the annotation models mostly depends on the time that annotators spend on creating annotations~\cite{zlateski2018importance}.
         
         By employing active learning strategy~\cite{6521405,xia2015accurate} and setting restrict constraints, those images that are difficult to be interpreted can be screened out and then manual annotated by experts. The received feedback can then be used to purify the annotation model through a loop learning way. Consequently, a large number of annotated images can be acquired to optimize the interpretation model and further boost the annotation task in an iterative way. With the iteration process, the number of images to be annotated will be greatly reduced to relieve annotation labor. The general workflow of semi-automatic image annotation is shown in Figure~\ref{figure:semi_annotation}. Benefiting from the excellent feature learning ability, deep learning based methods can be developed for image annotation with significant improvement of quality and efficiency~\cite{andriluka2018fluid}. 
         Instead of annotating the full image, human intervention by simple operations, \eg, point-clicks~\cite{chen2018tap}, boxes~\cite{papadopoulos2017extreme}, and scribbles~\cite{lin2016scribblesup}, can significantly improve the efficiency of interactive annotation. By utilizing the semi-automatic annotation strategy, a large-scale annotated RS image dataset can be constructed efficiently and also with quality assurance owing to the involvement of human labor. 
      \end{itemize}
        
        \subsubsection{Quality Assurance} 
       The dataset with high annotation quality is important for developing and evaluating interpretation algorithms. The following introduced strategies can be employed for the quality control when creating a dataset for RS image interpretation. 
       \begin{itemize}
       	\item {\bf Rules and Samples}
       The annotation rules without ambiguity are the guarantee of creating a high-quality dataset. Specifically, annotation rules like category definition, annotation format, viewpoint, occlusion, image quality, and others should be explained clearly. For example, whether to exclude the objects in occlusion, whether to annotate the objects of small sizes. If there are no clear rule descriptions, different annotators will annotate the image with their individual preferences~\cite{PASCALVOC}. For annotation in RS images, it is difficult for annotators to recognize the categories of ground features if they have no professional backgrounds. Therefore, samples are better to be provided by experts in the field of RS image interpretation and then presented to annotators as references. 
       
       \item{\bf Training of Annotators}
       Each annotator is required to pass the test of annotation training. Specifically, each annotator is given a small part of the data and asked to annotate the data to meet the articulated requirements. Those annotators that failed to pass the test cannot be invited to participate in the later annotation project. With such a design, dataset builders are able to build an excellent annotation team. Take xView~\cite{xview} as an example, the annotation accuracy of objects is vastly improved with trained annotators. Therefore, the training of annotators can be a reliable guarantee for high-quality image dataset annotation. 
       
       \item {\bf Multi-stage Pipeline}
       A serial of different annotation operations are easy to cause fatigue and result in annotation errors. To avoid this problem, the pipeline of multi-stage annotation can be designed to decouple the difficulties of the annotation task. For example, the annotation of object detection can be decoupled to be spotting, super-category and sub-category recognition~\cite{COCO}. By this method, each annotator only needs to focus on one simple stage during the whole annotation project and the error rate can be effectively decreased.
       
       \item{\bf Grading and Reward}
       A comprehensive evaluation of annotators can be performed with the annotation result. For example, the analysis of an annotators' behavior, \eg, the required time per annotation stage and the amount of annotation result over a period, can be conducted to assess the potentially weak annotations. Thus, different types of annotators can be identified, \eg, spammers, sloppy, incompetent, competent and diligent annotators~\cite{kazai2011worker}. Then, incentive mechanism (\eg, financial payment) can be employed to reward the excellent annotators and eliminate the inferior labels from unreliable annotators.
       
       \item {\bf Multiple Annotations}
       A feasible measurement to guarantee high-quality image annotation is to obtain multiple annotations from different annotators, merge the annotations and then utilize the response contained in the majority of annotations~\cite{Imagenet}. To acquire high-quality annotations, majority voting can be utilized to merge multiple accurate annotations~\cite{vijayanarasimhan2014large}. One disadvantage of this approach is that multiple annotations require more annotators and it is not reliable if the majority of annotators produce low-quality annotations.

       \item{\bf Annotation Review}
       Another effective method to ensure the annotation quality is to introduce the review strategy, which is usually integrated among other annotation pipelines when creating a large-scale image dataset~\cite{LabelMe}. Specifically, some annotators can be invited to conduct peer review and rate the quality of the created annotations. Besides, further review work can be conducted by experts with professional knowledge. Based on the reviews of supervisors in each annotation step, the overall annotation quality can be strictly controlled in the whole annotation process. 
       
      \item{\bf Spot Check and Assessment}
      To check the annotation quality, a test set can be sampled from the annotated images. Also, gold data can be created by sampling and labeling a proper proportion of images annotated by experts. Then, one or several interpretation models can be trained based on these datasets and the interpretation performance (\eg, $Recall$ and $Precision$ for object detection~\cite{DOTA,ding2019learning}) can be evaluated to compare annotation from annotators and gold data from experts. If the evaluation result is lower than the preset expectation, annotations from the corresponding annotator would be rejected and required to be resubmitted for repetitive annotation. 
      \end{itemize}

\section{An Example: Million-AID} 
\label{ExampleMAID}
 Following the aforementioned prototype for building  benchmark datasets for RS image interpretation, in this section we present an example to construct a large-scale benchmark dataset for RS scene classification, \ie, the {\em Million Aerial Image Dataset} (Million-AID). Limited by the scale of scene images and number of scene categories, current datasets for scene classification are far from meeting the requirements of the real-world feature representation and the scale for interpretation model development. It is desperately expected that there is a much reliable dataset for scene classification in RS community. In this section, we build Million-AID in the spirit of~{\em \textbf{DiRS}}. And the introduced coordinates collection strategy is employed for efficient scene image acquisition. The dataset quality is guaranteed with a handful of human labor, which finally formulates a semi-automatic and reproducible framework for the construction of RS image scene dataset. The constructed Million-AID will be released for public accessibility. 

    \begin{figure*}[t!]
      \centering
      \includegraphics[width=0.82\linewidth]
      {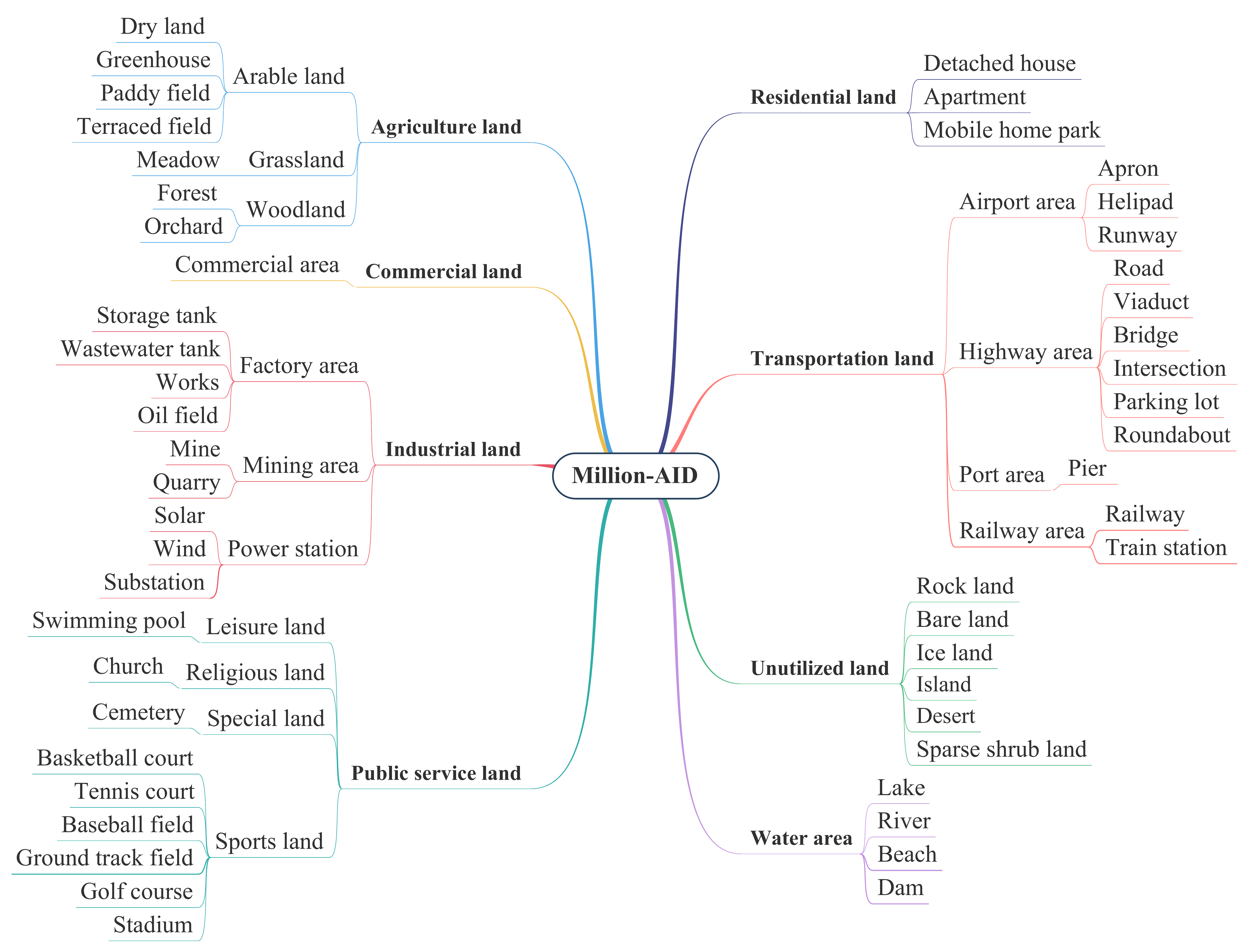}
      \caption{The hierarchical scene category network of Million-AID. All categories are hierarchically organized in a three-level tree: 51 leaf nodes fall into 28 parent nodes at the second level which are grouped into 8 nodes at the first level, representing the 8 underlying scene categories of agriculture land, commercial land, industrial land, public service land, residential land, transportation land, unutilized land, and water area.}
      \label{figure:MAID_ClsNet}
      \vspace{-3mm}
    \end{figure*}
  
    \subsection{Scene Category Organization}
      \subsubsection{Main Challenges in Application}
       Benefiting from the advancement of RS technologies, the accessibility of RS images has been greatly improved. However, the construction of a large-scale scene classification dataset still faces challenges in aspects like scene taxonomy and image diversity. Obviously, a complete taxonomy of RS image scenes is better to have wide coverage of categorical space since there are a large number of semantic categories in practical applications, \eg, LULC. With various scene images in different categories, the completeness of a scene taxonomy is also significant to enhance the diversity of the dataset. Thus, the determination of scene categories is of great significance to construct a high-quality and practical RS image dataset for scene classification. Some existing datasets, such as the UCM~\cite{UCM}, RSSCN7~\cite{RSSCN7}, and RSC11~\cite{RSC11}, contain limited scene categories, which make the them not sufficiently represent the diverse content reflected by RS images. Consequently, the scene classification models learned from datasets of limited categories usually show weak generalization ability.
      
      When facing practical applications, the excellent organization of scene categories is an important feature for scalability and continuous availability of a large-scale RS image dataset. Typically, the semantic categories which are closely related to human activities and land utilization are selected for the construction of scene categories. Because of the complexity of RS image content, there is a large number of semantic categories and also a hierarchical relationship among different scene categories. Usually, it is difficult to completely cover all the semantic categories and the relationship information between different scene categories can be easily neglected, owing to the subjectivity of dataset builders. Therefore, effective organization of scene categories should be of great significance to construct a RS image dataset of high quality and scalability. 
    
    \subsubsection{Scene Category Network}
      Faced with the above challenges, we build a hierarchical network to manage the categories of RS image scenes, as shown in Figure~\ref{figure:MAID_ClsNet}. To satisfy the requirements of practical application rather than the classification algorithms, we construct the scene category system by referencing to the land-use classification standards of China (GB/T 21010-2017). Considering the inclusion relationships and content discrepancies of different scene categories, the hierarchical category network is finally built with three semantic layers. In accordance with the semantic similarity, those categories with overlapping relationships are merged into a unique semantic category branch. Thus, the scene classification dataset can be constructed with category independence and semantic completeness. 

      As shown in Figure~\ref{figure:MAID_ClsNet}, the proposed category network is established upon a multi-layered structure, which provides scene category organization with different semantic levels. When it comes to the specific categories, we extract aerial images on Google Earth and determine whether the images can be assigned with the semantic scene labels in the category network. For those images that cannot be recognized with specific categories within the existing nodes, new category nodes will be embedded into the original category network by experts according to the image scene content. In view of the fact that there are inclusion relationship among different scene categories, all classes are hierarchically arranged in a three-level tree: 51 leaf nodes fall into 28 parent nodes at the second level, and the 28 parent nodes are grouped into 8 nodes at the first level, representing the 8 underlying scene categories of agriculture land, commercial land, industrial land, public service land, residential land, transportation land, unutilized land, and water area. Benefiting from the hierarchical structure of category network, the scene labels from the  parent nodes can be directly assigned to the images belonging to the corresponding leaf nodes. Therefore, each image will possess semantic labels with different category levels. This mechanic also provides potentiality for scene classification at flexible category levels. 

      As can be seen, the category definition and organization can be achieved by the proposed hierarchical category network. The synonyms of the category network are relevant to the practical application of LULC and hardly need to be purified. One of the most prominent advantages of the category network lies in its semantic structure, \ie, its ontology of concepts. Hence, a new scene category can be easily embedded into the constructed category network as a new branch of synonym. The established category hierarchy can not only serve as the category standard for Million-AID dataset but also provides a valuable reference for dataset construction toward other interpretation tasks. Thus, these properties endow our proposed dataset with high practicability when facing real applications. 
      
     \begin{figure*}[htb!]
        \centering
        \includegraphics[width=0.85\linewidth]
        {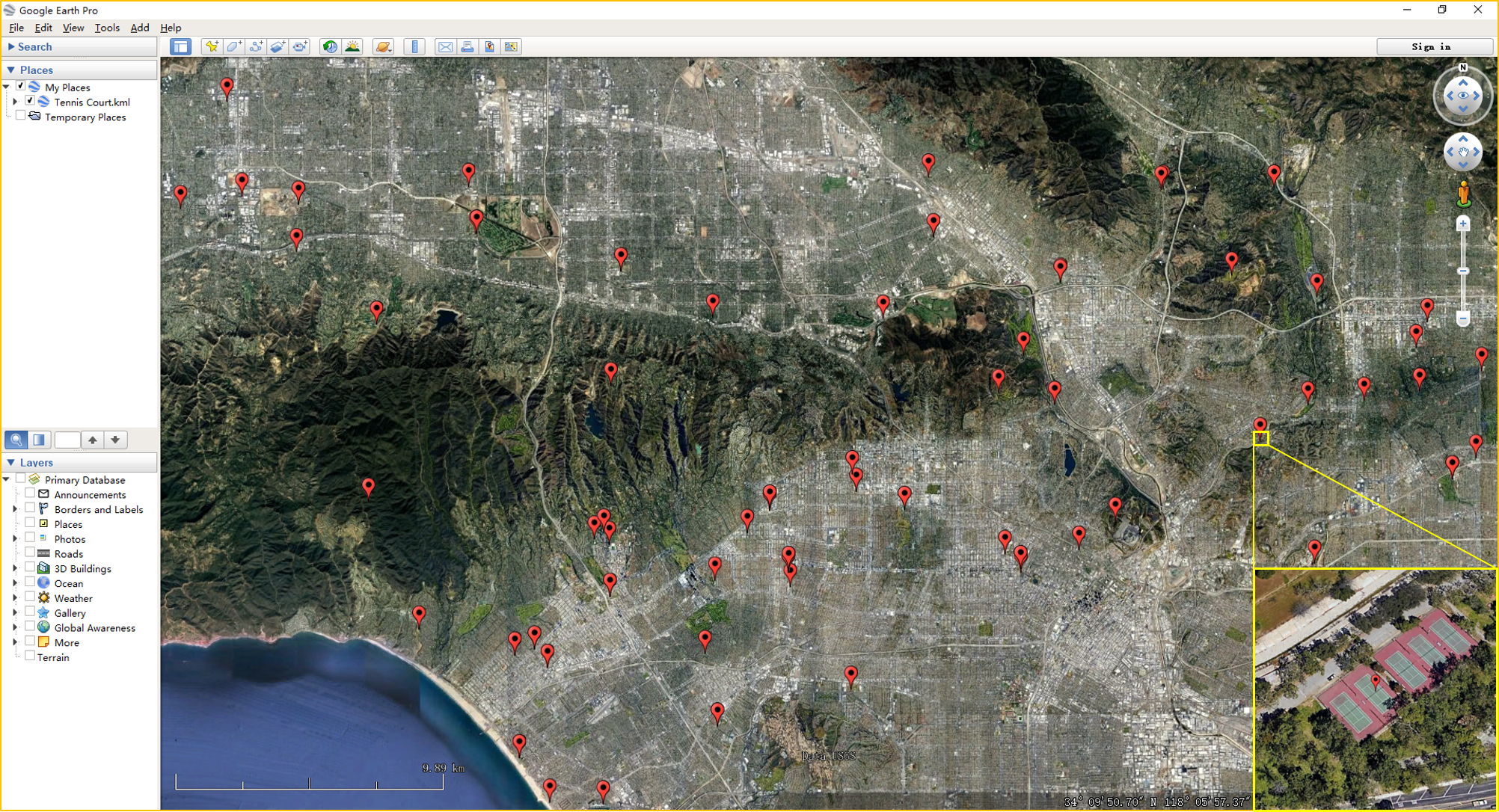}
        \caption{The points of searched tennis courts shown in Google Earth Pro~({\em \copyright{2020 Google LLC.}}), where the top-left and bottom-right coordinates are ($\ang{34.1071}$ N, $\ang{118.3605}$ W) and ($\ang{33.9823}$ N, $\ang{118.3605}$ W), respectively. We consider the tennis courts as point ground features. The red marks show the searched locations of tennis courts. The eagle window shows the detail of a tennis court scene, which confirms the validity of collecting semantic coordinates by our proposed method.} 
        \label{figure:TennisCourtSearch}
      \end{figure*}

   \subsection{Semantic Coordinates Collection}
     In the conventional pipeline of constructing a scene classification dataset, one needs to manually search the target region that contain specific scenes. Then the scene images are collected from the image database. However, finding the target region with given semantic scenes is a time-consuming procedure and usually requires high-level technical expertise. Besides, in order to ensure the reliability of scene information, images need to be labeled by specialists with domain knowledge of RS image interpretation. To alleviate this problem, we employ the introduced coordinates collection strategy and interactive annotation methodology to build the scene classification dataset. Specifically, we employ public map search engines, open sourced data, and public geodatabase resources to collect and label RS scene images. With the rapid development of geographic information and RS technologies, there are rich and publicly available geographic data like online map, open source data, and archives published by agencies as introduced before. Typically, these public geographic data present the surface features in forms like point, line, and plane, which describe the semantic information of ground objects and carry corresponding geographic location information. Based on the public geographic data, we search for coordinates of specific semantic tags, and then utilize the semantic coordinates to collect the corresponding scene images. 
     
     In RS images, scenes are presented with different geometric appearances. In the case of our practice, different methods are presented to acquire the labeling data. Google Map API and publicly available geographic data are mainly employed to obtain the coordinates of point features while OSM API is mainly utilized to acquire the coordinates of line and plane features. In application, these methods can be combined to obtain different coordinate data of different forms. The acquired coordinates are then integrated into block data which presents the scene extent. Finally, the block data are further processed to obtain scene images which are automatically assigned with scene labels.  

      \begin{figure*}[htb!]
        \centering
        \includegraphics[width=0.85\linewidth]
            {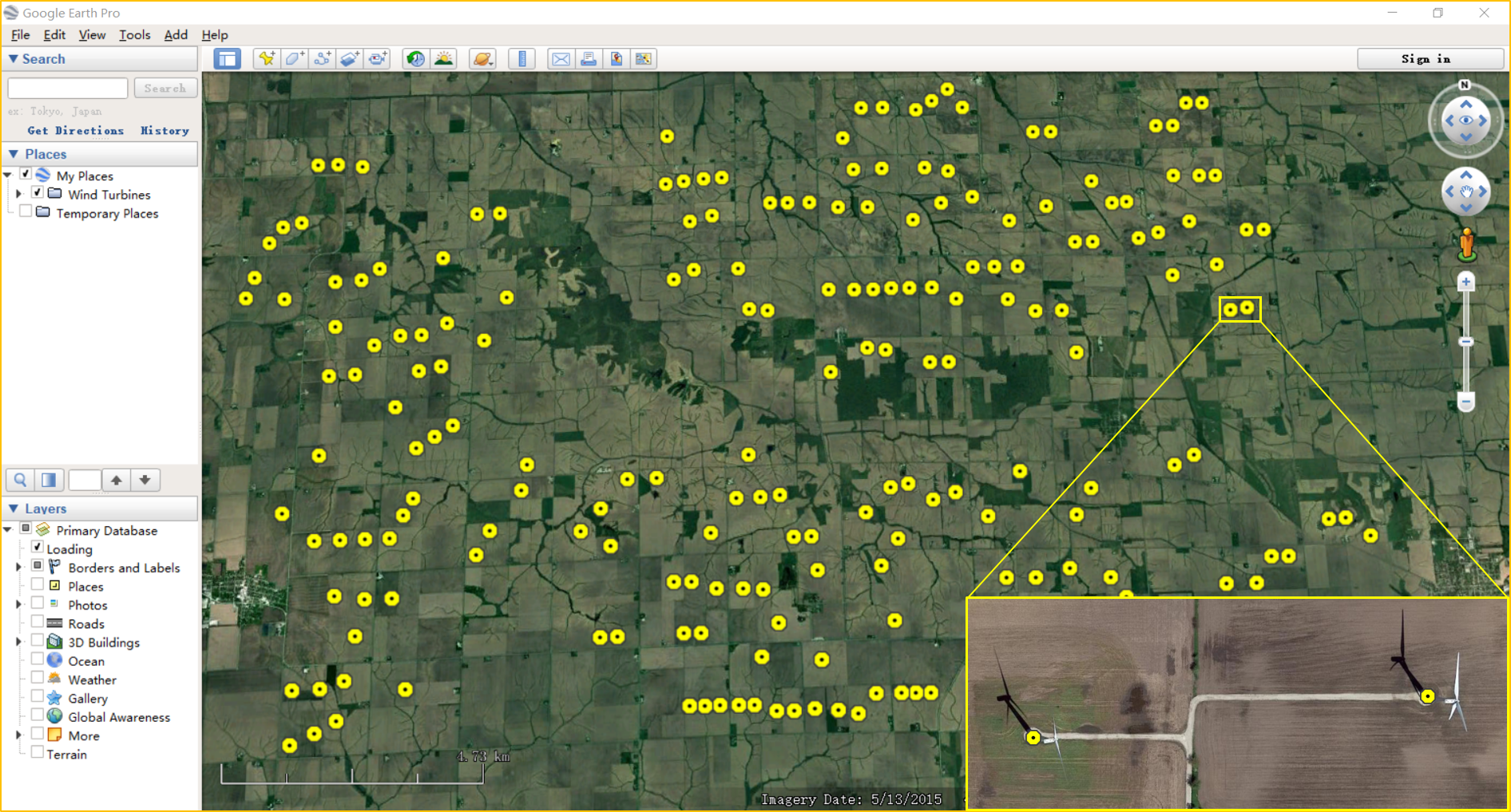}
         \caption{The points of wind turbines extracted from USWTDB and integrated in Google Earth Pro~({\em \copyright{2020 Google LLC.}}), where the geographic range is indicated with the top-left coordinates ($\ang{41.2695}$ N, $\ang{90.3315}$ W) and bottom-right coordinates ($\ang{41.1421}$ N, $\ang{90.0424}$ W). The eagle window shows the details of two wind turbines.}
        \label{figure:WindTurbineSearch}
        \vspace{-3mm}
      \end{figure*}
     
     \subsubsection{Point Coordinates}
     The point features, such as tennis courts, baseball fields, basketball courts, and wind turbines, take relatively small ground space in the real-world. The online Google map makes it possible to discover the world with rich location data, \eg, over 100 million places. This provides a powerful solution to search the ground objects with specific semantic tags. Therefore, we develop a semantic tag search tool based on the Google Map API. With the customization search tool, we input semantic tags to retrieve corresponding point objects using the online map search engine and obtain the geographic coordinates that match the semantic information within a certain range. The retrieved point results with location information, \ie, geographic coordinates, are naturally attached with scene tags. Figure~\ref{figure:TennisCourtSearch} shows the search result returned by the semantic tag ``baseball field'' based on the tool. To enhance the diversity of the dataset, we search points of interested objects through a wide range of geographic areas. This strategy makes it possible to cover individual point objects in distinct positions, which is able to greatly enhance the within-class diversity and quickly obtain a large number of points with semantic tags.

     The map search engines have provided a powerful interface for accessing point data. However, many of them are associated with categories of common scenes, which will limit the diversity of dataset. For those scenes related to specific scene categories, it is reasonable to employ the publicly available geographic information and obtain the point data. Using the online platforms that publish geographic dataset, we collect the coordinate data of storage tanks, bridges, and wind turbines. For example, the U.S. Wind Turbine Database (USWTDB)\footnote{https://eerscmap.usgs.gov/uswtdb} provides a large number of locations of land-based and offshore wind turbines in the United States. Figure~\ref{figure:WindTurbineSearch} shows the tagged data of wind turbines, which can indicate the accurate positions of wind turbines in a local area. By processing these data, a single point coordinate data corresponding to a wind turbine scene is obtained. Consequently, with the publicly released geographic information, we are able to employ the strategy of geodatabase integration to collect coordinates of specific scenes.

     \begin{figure*}[t!]
        \centering
        \includegraphics[width=0.87\linewidth]
            {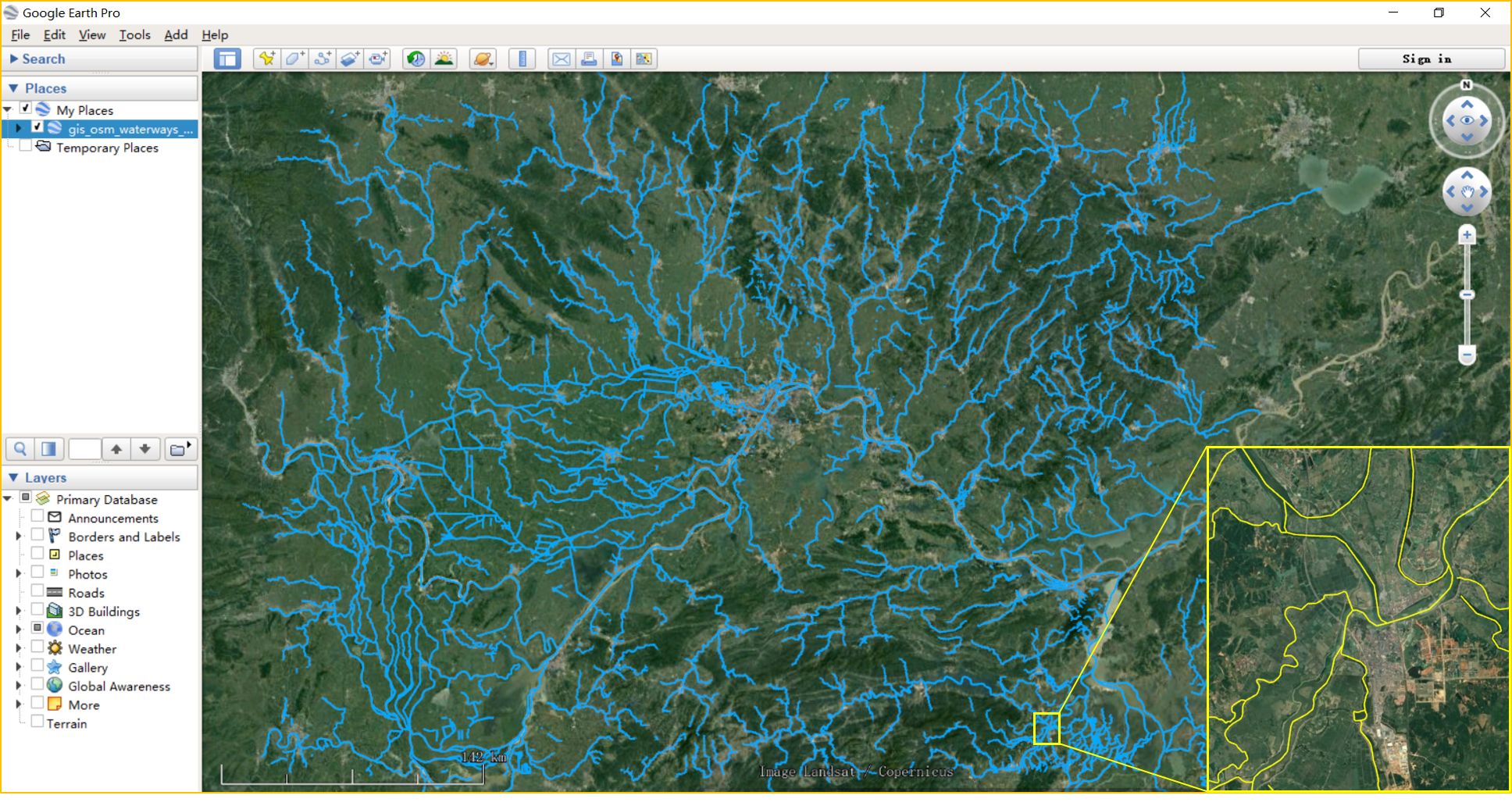}
        \caption{The river lines within a local area of China are extracted from OSM and displayed in Google Earth Pro~({\em \copyright{2020 Google LLC.}}), where the upper left and bottom right coordinates of geographic range are ($\ang{32.2826}$ N, $\ang{111.1027}$ E) and ($\ang{28.7477}$ N, $\ang{118.5372}$ E), respectively. The zoomed image shows the details of river lines.}
        \label{figure:RiverLine}
        \vspace{-3mm}
     \end{figure*}
      
      \subsubsection{Line and Plane Coordinates}
      The ground features, such as river and railway, are usually presented in the form of lines. Other features like grassland and residential land are typically presented by planes. In order to obtain the scene images of line and plane features, we employ the open source data for scene coordinates collection as introduced before. Specifically, the OSM is utilized to extract the location information of line and plane features. OSM is a collaborative project to create a free editable map of the world. The elements in OSM consist of node, way, and relation, which are also the basic components of OSM conceptual data model that depicts the physical world. A node represents a point feature on the ground surface. It can be defined by a pair values of latitude and longitude. The way feature is composed of two or more connected nodes. An open way describes a linear feature, such as roads, streams, and railway lines. A plane or area feature can be described in a closed way. A relation element in OSM is utilized to describe one or several complex objects with a data structure that records a relationship between nodes, ways, and other relations. Every node or way has tags and geographic information that describe the corresponding ground object. Therefore, a line or plane feature that belongs to certain semantic classes can be obtained by searching its corresponding tags. 
      
\begin{figure*}[t!]
        \setlength{\abovecaptionskip}{-0.1cm}
        \centering
        \includegraphics[width=0.85\linewidth]
            {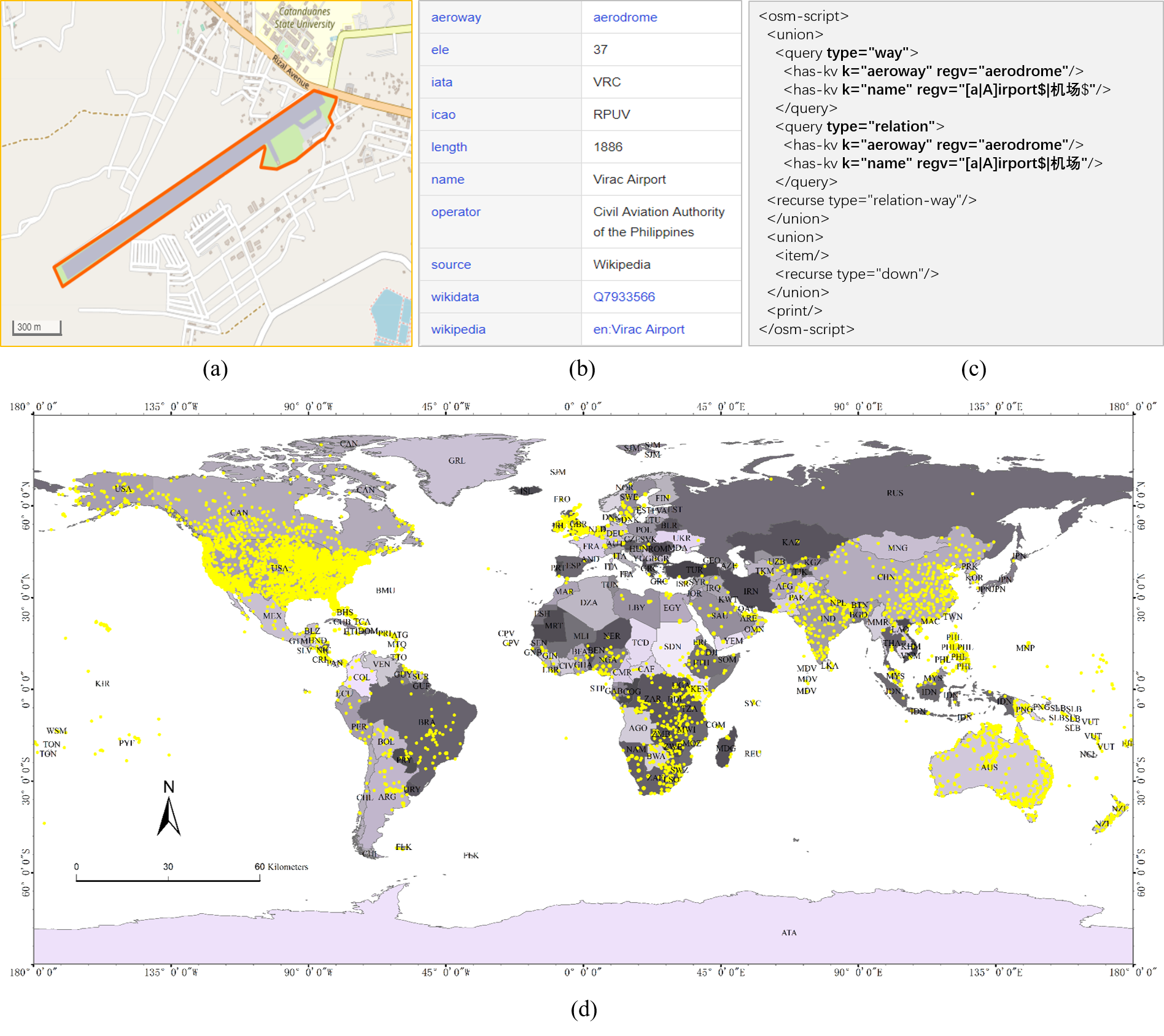}
        \caption{The illustration of searching scenes of airports around the world. An airport in OSM contains a large amount of tags, which can be employed to search airports with specific semantic key-value labels, \eg, {\em aeroway} and {\em name}. More than 5,000 world airports in forms like way and relation can be obtained with accurate geographic coordinates, using English and Chinese semantic tags.}
        \label{figure:AirportSearch}
        \vspace{-3mm}
      \end{figure*}

\begin{figure*}[t!]
        \centering
        \includegraphics[width=0.87\linewidth]
            {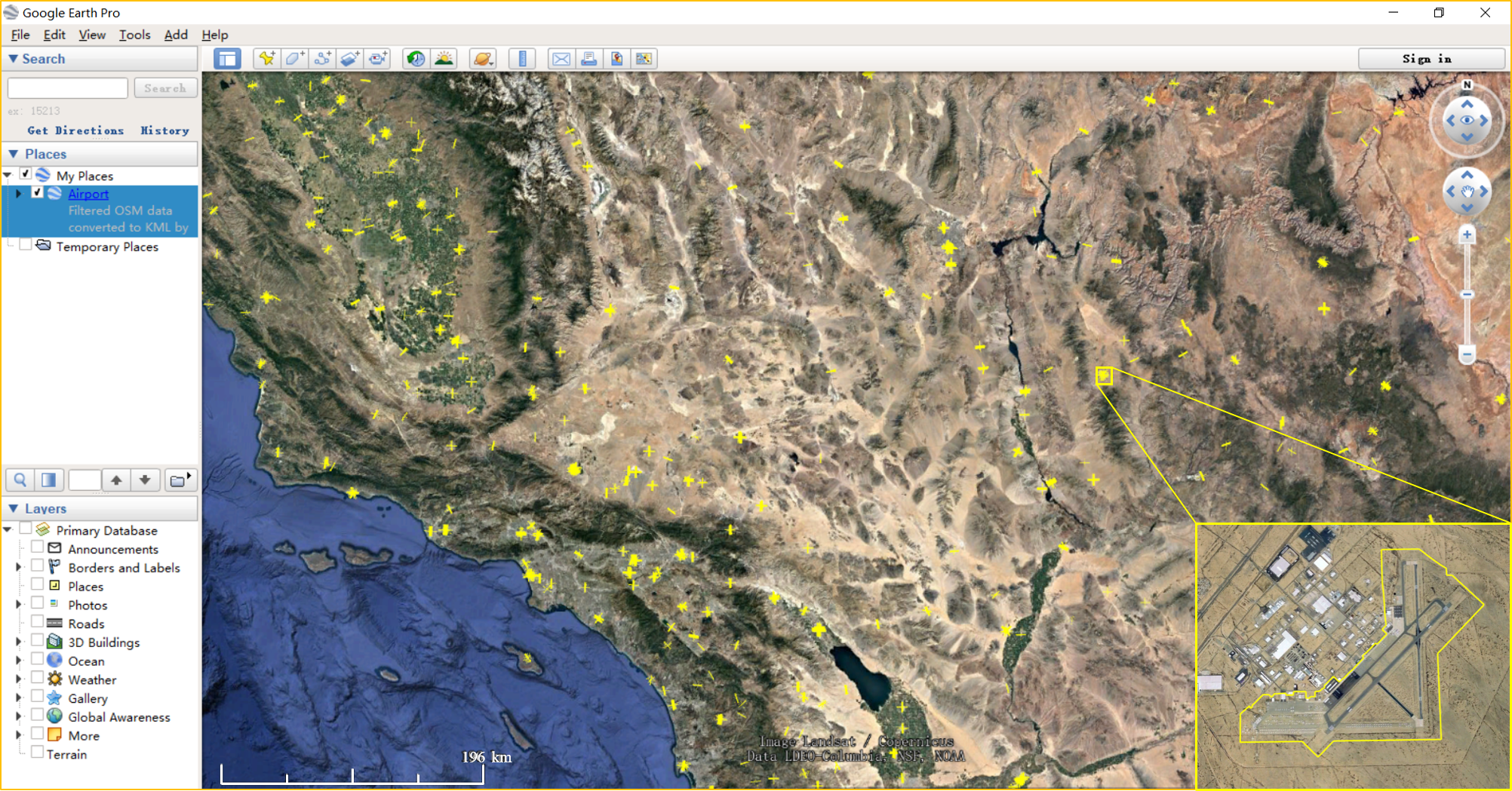}
        \caption{The airport planes within a local area of the United States are extracted from OSM and shown in Google Earth Pro~({\em \copyright{2020 Google LLC.}}), where the upper left and bottom right coordinates of geographic range are ($\ang{37.2262}$ N, $\ang{115.8819}$ W) and ($\ang{32.6005}$ N, $\ang{110.6497}$ W), respectively. The zoomed image shows the accurate extent of an airport in a closed OSM way feature.}
        \label{figure:AirportPlane}
        \vspace{-3mm}
      \end{figure*}

      Many methods can be employed to obtain the geographic coordinates of ground features from OSM. As the most convenient way, we collect the line and plane features directly from the free, community-maintained data, \eg, Geofabrik\footnote{http://www.geofabrik.de/geofabrik}, produced by OSM. Figure~\ref{figure:RiverLine} shows the river line features collected through Geofabrik, which provides maps and geographic data extracted from OSM. Besides, we also employ the OSM interface, \ie, Overpass API, to extract the features of interest. In order to obtain the semantic coordinates of scenes in the constructed network, we also search features by utilizing the powerful query language. The query criteria are associated with location information, type of objects, proximity of tag properties, and their combinations. Figure~\ref{figure:AirportSearch} shows the illustration of searching scenes of airport areas around the world. And the searched airports within a local area of the United States are integrated into Google Earth as shown in figure~\ref{figure:AirportPlane}. It can be seen from Figure~\ref{figure:AirportSearch}-\ref{figure:AirportPlane} that the extracted plane data is consistent with the real-world airport scenes, and thus, the semantic label is reliable. These results indicate that the former introduced method of employing the open source data is a practical, efficient, and reliable strategy for scene image acquisition via the collection of semantic scene coordinates. 
      
    \subsection{Scene Image Acquisition}
\begin{figure*}[t!]
        \setlength{\abovecaptionskip}{-0.05cm}
        \centering
        \includegraphics[width=0.85\linewidth]
        {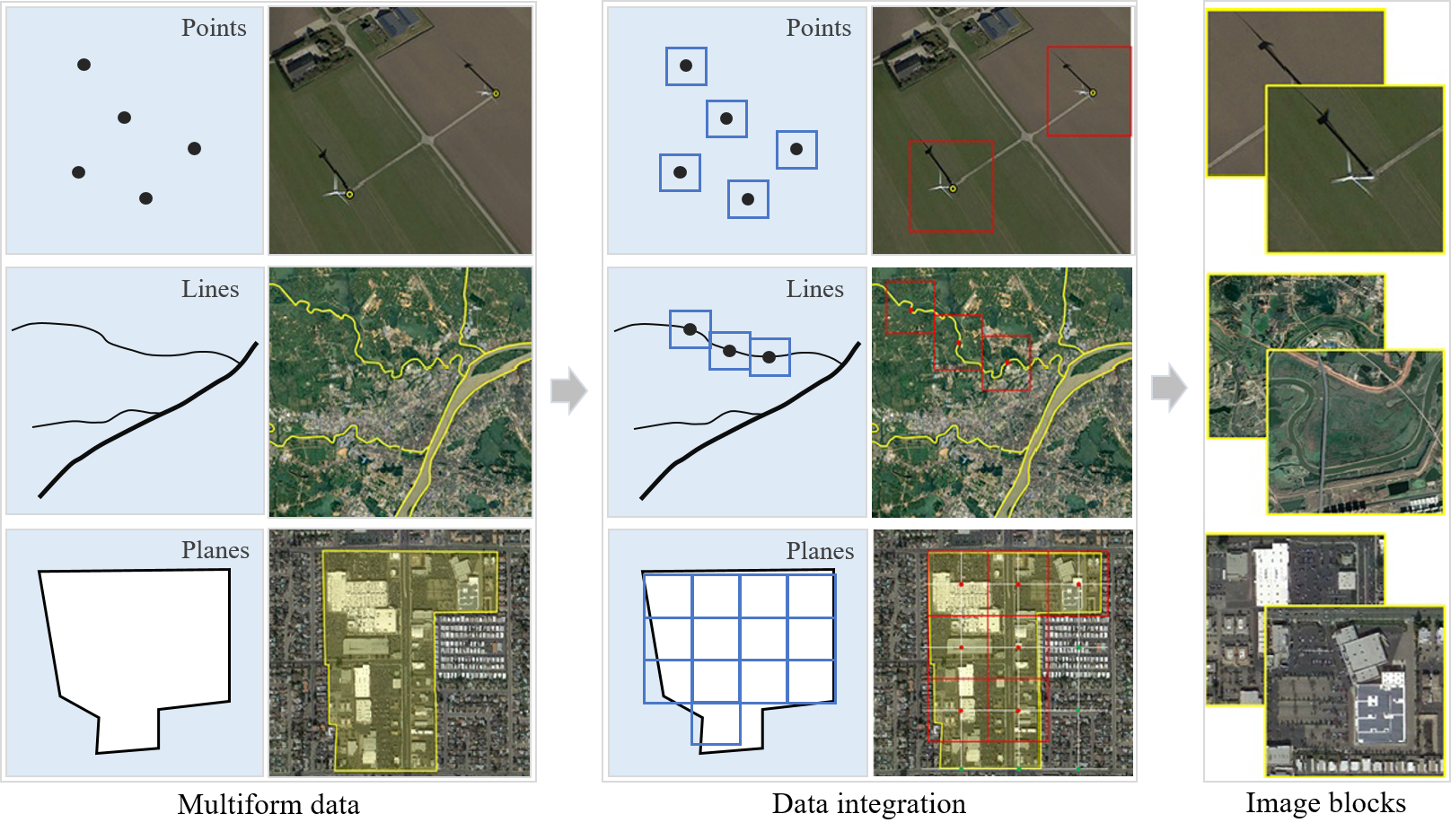} 
        \caption{The illustration of the acquisition of RS scene images based on the collected geographic point, line and area data. The points are set as the centers of scene square boxes. For line data, the center points are sampled by intervals. For plane data, scene square boxes are generated by mesh grids. The red frames indicate the generated scene square boxes which is consistent with the final scene image blocks.}
        \label{figure:AIDconstruct}
        \vspace{-3mm}
     \end{figure*}

    The geographic point, line, and plane coordinates collected through the above processes are employed to extract scene images from Google Earth. Figure~\ref{figure:AIDconstruct} illustrates the overall framework of collecting RS scene images. For the searched point data, the coordinates are attached with specific tags of semantic categories and we take the geographic coordinates as the center of a square box. For line data, we sample the points along the line by intervals and a sampled point is selected as the center of a square box. Based on the center point, a square box of customized size is generated to serve as a scene box characterized by four geographic coordinates. For the plane data, \eg, commercial area, a mesh grid is generated to divide the plane area into individual scene boxes. Some scenes like airport and train station are usually presented with individual blocks. Therefore, the envelop rectangles of these blocks are extracted as the scene boxes directly. Thus, the content of a scene box is consistent with its corresponding scene category. 
    
    All the scene boxes are utilized to outline and download scene images from Google Earth. The scene images are extracted with different sizes, such as $256\times256$ and $512\times512$, according to the scene scales and resolutions of Google Earth images. There may be inaccurate semantic label assignments caused by noisy coordinates and wrong scene boxes. To ensure the correctness of the category labels, all of the scene images are checked by specialists in the field of RS image interpretation. Specifically, those downloaded images in a specific category are deleted if they are assigned with wrong scene labels. For those scene boxes that are overlapped with each other, only one of the scene boxes will be chosen to extract the corresponding scene image. With these operations, we are able to improve the accuracy of the  scene images that are automatically annotated, and therefore, guarantee the quality of the constructed datasets for scene classification. 
    
\subsection{Discussion}
    By following the presented guidances, the above procedure formulates a framework for the collection of RS scene images. As shown in Table~\ref{tab:SCdataset}, Million-AID consists of the most scene categories compared to the existing scene classification datasets except for fMoW~\cite{fWoM}. Different form the existing datasets of which categories are organized with parallel or uncertain relationships, scene categories in Million-AID are organized with systematic relationship architecture, giving it superiority in management and scalability. More importantly, the scene categories are customized to match the land-use classification standards. All of these have greatly enhanced the practicability of the constructed dataset.
    
    % diversity
    The property of diversity is important for a dataset to train interpretation algorithms of strong generalization ability. In the construction of Million-AID, the diversity of scenes in each category is greatly enhanced by the wide geographic distribution of scene locations. Specifically, images in each scene category are extracted from different areas around the world. When some acquired scene coordinates are intensively located in a local area, we try to collect more scene coordinates from other different areas to increase the scene diversity in spatial distribution. The advantage of this strategy is obvious as the wide distribution of scene coordinates makes our collected scenes of interest be individually and spatially independent. Each scene object is able to reflect the unique characteristic from different perspectives. Thus, the within-class diversity of scene images can be greatly improved. In addition, the large-scale semantic categories also improve the between-class diversity of scene images. Some scenes belonging to different categories may share similar characteristics,~\eg, stadium and ground track field. Especially, the fine-grained scene categories also possess the same semantic information as they all belong to the same parent scene class. In general, the independent scene objects within the same category and the unique characteristics among different scene categories make the constructed dataset characterized with big diversity.

    As introduced before, the collected images in Million-AID are mainly from Google Earth. It is well-known that images in Google Earth are from different satellites, including but not limited to the GeoEye, WorldView, QuickBird, IKONOS, SPOT, and Landsat serial satellites. The multiple data sources can naturally improve the richness of the scene images. Besides, scene images in Million-AID are with broad resolutions, ranging from 0.5m to 153m per pixel. Note that there are also images of different resolutions in the same scene category, in which scene images are acquired according to the scene scale. At the same time, in the process of scene image inspection, we pay more attention to choose scene images under different imaging conditions,~\eg, viewpoint and illumination, to increase the richness of scene images. In the scene image acquisition stage, those scene images with regional overlap are eliminated and only one of them is retained. Thus, the collected content information of each scene image is not repeated and different scene images have different scene background information. In this way, the variety of dataset is greatly guaranteed by the large-scale and individual scene images. These characteristics allow us to greatly enhance the richness of the dataset at the image level.

    % annotation 
    The construction process of Million-AID also largely follows the idea of semi-automatic annotation. At the stage of scene location acquisition, map search engines, open source data, and public geographic information database are used to obtain the point, line, and plane features that indicate the scene objects. The scene labels are then acquired with the corresponding semantic tags of coordinates. In practice, each kind of scene objects can be acquired by combing several scene coordinate collection methods as introduced. Compared with the manual search method, our method can greatly reduce the difficulty of scene information acquisition and improve the efficiency of dataset construction. Thus, it is easy to collect large-scale scene images. Consequently, the number of images in each scene category goes beyond 2,000 and reach over 20, 000 in average. All of these provide guarantees of diversity and richness for the constructed dataset. Not only that, owing to our strategy that automatically obtain scene coordinates and semantic annotations, only image check and deletion are performed manually, which is a really easy work. Therefore, the interactive annotation strategy makes the dataset construction falls into a semi-automatic annotation mode which can greatly reduce the manpower cost and ensure the label quality simultaneously. Thus, it is feasible to build the large-scale scene image dataset with high quality. Consequently, following the introduced guidances and annotation methodology, the Million-AID dataset is achieved with more than 1,000,000 annotated semantic images of 51 scene categories.

\section{Challenges and Perspectives} \label{Resources}
   Driven by the wide applications of RS image interpretation, various datasets have been constructed for the development of interpretation algorithms. In spite of the great success in RS image datasets construction over the past years, there are still a giant gap between the requirement of large-scale dataset and interpretation algorithm development, especially for data-driven methods. Thus, how to speed up the annotation process of RS images remains to be a key issue for the construction of interpretation datasets. After investigating the current annotation strategies for RS image datasets, this section discusses the current challenges and potential perspectives for efficient dataset construction. 

  \subsection{Visualization Technology for RS Image Annotation }
  In the process of RS image annotation, semantic content in the image is firstly recognized by visual interpretation of experts. Then, the semantic labels are assigned to the corresponding objects in pixel, region, or image levels. Thus, the visualization technology for RS image plays a significant role in the process of accurate semantic annotation, especially for the hyper-spectral, SAR, and large size RS images. 
 
    {\em Hyper-spectral image annotation with visualization technology.} 
    A hyper-spectral image usually contains hundreds of spectral bands, which can provide rich spatial-spectral information of features of the Earth's surface. However, the high dimensionality of hyper-spectral image brings the challenge for semantic information annotation. The reality is that the existing display devices are designed for gray or color images with typical RGB channels. Thus, it is impossible to directly display a hyper-spectral RS image which consists of hundreds of spectral bands using conventional display strategies. In order to alleviate this problem, the strategy of band selection can be explored to choose three representative bands of the original image as RGB channels~\cite{KANG2020130}. The fundamental idea of this strategy is to select bands with as much information as possible from the original hyper-spectral image or directly reference the characteristics of the annotation objects. Alternatively, band transformation can also be considered by making the best use of the rich bands. The basic principle is to transform the original image into a new feature space by spectral transformation, \eg, dimensionality reduction, band fusion, and clustering. Then, the three representative channels can be selected for visualization~\cite{8907846}. These strategies should rely on effective algorithms developed for band selection and transformation. Besides, hyper-spectral RS images collected from different sensors usually suffers from spectral variability, making it difficult for information extraction and content annotation. Thus, effective hyper-spectral unmixing method~\cite{hong2018augmented} can be developed to accurately estimate the content to be annotated. Thus, hyper-spectral RS images can be well visualized, providing a guarantee for annotating reliable information. 
    
    {\em SAR image annotation via physical signal expression.}
    Compared with optical RS images, the challenge of SAR image annotation mainly comes from the weak legibility in visual appearance. SAR’s all-weather, all-day, and penetrating imaging ability endow it with great superiority over optical RS images in some practical applications, \eg, disaster rescue. However, due to the interference of coherent returns scattered by small reflectors within each resolution cell, SAR images are contaminated by multiplicative noise called speckle. Also, SAR images are usually with gray-scale mode and there is not any color information except for full-polarimetric SAR images~\cite{8985381}. All of these pose great challenges for SAR image annotation. An essential point is that the SAR image is represented with signal information, where different objects show different polarimetric features. Thus, the utilization of physical information of objects can be a promising solution for SAR image annotation, relying upon the basic principles related to surface roughness/smoothness and changes in the back-scattering signal intensity of surface conditions~\cite{6132463}. On the other hand, visualization technology should also be explored to enhance the legibility of SAR image content. One direction is to colorize the non-full-polarimetric to full-polarimetric SAR images based on the radar polarimetry theories. Inspired by the success of transfer learning in computer vision, it is also valuable to color the SAR images through simulating RGB images using DCNNs~\cite{8985381}. With these considerations, more efforts should be poured into SAR image visualization to relieve the difficulties of annotation.

    {\em Large-size image annotation with high interaction efficiency.} 
    Annotation for large size RS images is another important challenge. Currently, RS images are usually annotated by tools designed for natural image labeling~\cite{LabelMe,LabelImg}, where only images of limited sizes, \eg, image width/height of several hundred pixels, can be fully visualized for interactive annotation. However, with the improvement of image resolution, RS images taken from the bird-view have large geographic coverage and thus possess large sizes,~\eg, width/height of tens of thousands pixels. Thus, the conventional annotation solution can only visualize the local region of a RS image for annotation operations. Besides, current machine monitor devices are also with limited sizes and resolution. It requires constant image roaming and zooming operations when annotating large size RS images, which heavily hinders the interaction efficiency of annotation and loss the possibility of catching the features with spatial continuity from a global perspective of the image content. On the other hand, the RS images with spatial information needs large storage space because of its large amount of data. Thus, the visualization of RS images also requires large-scale computing capability when conduct annotating operations. Considering these points, the visualization technology for displaying, roaming, zooming, and annotating large size RS images needs to be stressed for efficient annotation.

    \newcommand{\tabincell}[2]{\begin{tabular}{@{}#1@{}}#2\end{tabular}}
    \renewcommand\arraystretch{1.6}
    \begin{table*}[htb!]
    \centering
    	\caption{Annotation tools for image dataset construction}
    	\vspace{-1mm}
        \begin{threeparttable}
        \begin{tabular}{c|m{0.16\textwidth}<{\centering}|c|c|m{0.5\textwidth}<{\centering}}
    		\hline
    		\textbf{No.}  &Name  &Ref.  &Year  &Description\\
    		\hline
    		1  &LabelMe  &\cite{LabelMe}  &2008  & An online image annotation tool that supports various annotation primitives, including polygon, rectangle, circle, line and point. \\
    		\hline
    		2  &Video Annotation Tool from Irvine, California (VATIC)  &\cite{VATIC}  &2012  &An online tool that efficiently scaling up video annotation with crowdsourced marketplaces (\eg, AMT). \\
    		\hline
    		3  &LabelImg  &\cite{LabelImg}  &2015  &A popular graphical image annotation application that labels objects in images with bounding boxes.\\
    		\hline
    		4  &Visual Object \qquad\qquad Tagging Tool \qquad (VOTT)  &\cite{VOTT}  &2017  &An open source annotation and labeling tool for image and video assets, extensible for importing/exporting data to local or cloud storage providers, including Azure Blob Storage and Bing Image Search. \\
    		\hline
    		5  &omputer Vision \qquad\qquad Annotation Tool \qquad\qquad  (CVAT)  &\cite{CVAT}  &2018  &A universal data annotation approach for both individuals and teams, supporting large-scale semantic annotation for scene classification, object detection and image segmentation. \\
    		\hline
    		6  &Image Tagger  &\cite{imagetagger}  &2018  &An open source online platform to create and manage image data and diverse labels (\eg, bounding box, polygon, line and point), with friendly support for collaborative image labeling.\\
    		\hline
    		7  &Polygon RNN++ &\cite{PolygonRNN++}  &2018  &A deep learning-based annotation strategy, producing polygonal annotation of objects segmentation interactively using humans-in-the-loop. \\
    		\hline
    		8  &Makesence.AI &\cite{makesenseai}  &2019  &An open source and online image annotation platform, using different artificial model to give recommendations as well as automate repetitive and tedious labeling activities.\\
    		\hline
    		9  &\tabincell{c}{VGG Image Annotator \\ (VIA)} &\cite{VIA}  &2019  &A simple and standalone manual annotation software for image and video, providing rich labels like point, line, polygon as well as circle and ellipse without project management.\\
    		\hline
    	\end{tabular}
    	\begin{tablenotes}
    	   \scriptsize
            \item [*] This table non-exhaustively presents the popular and representative image annotation tools. 
    	\end{tablenotes}
    	\end{threeparttable}
    	\label{tab:AnnotationTools}
    	\vspace{-3mm}
    \end{table*} 

  \subsection{Annotation Efficiency and Quality Improvement}
  There is no doubt that the constructed RS image dataset is ultimately utilized for various interpretation applications. Thus, the application products can be employed in turn to facilitate the annotation of RS images. And in the annotation process, the reliable tools developed for RS image annotation also play an important role in efficiency improvement. Besides, noise data is a common problem in RS image annotation, which makes the handling of noise annotation a valuable topic for dataset quality control as well as the development of interpretation algorithms.
  
   {\em Cooperation with application departments.} 
   A feasible way to improve the efficiency of RS image annotation is to cooperate with application departments and convert the application products to annotated datasets. Once the product data in the application department is produced, they naturally carry semantic information which can be utilized as the source of RS image annotations. For example, thematic map as the typical application product is able to be involved in creating training dataset and generating new annotation product~\cite{9121728,8635553,inglada2017operational}. Usually, the map data of land survey from the land-use institution is obtained through field investigation and thus possesses accurate land classification information, which can be easily combined with RS images to create reliable annotated datasets for model adaption. This scheme is reasonable because the product data from the application department is oriented to the real application scenarios. At this point the created dataset for RS image interpretation can most truly reflect the key challenges in the real application scenarios. Thus, the interpretation algorithms built upon this kind of dataset would be more practical. Besides, the product data will change with the alternation of application department's business. Thus, the product data can be employed to update the created dataset promptly. In this way, it ensures the established dataset to be always oriented to real applications, and therefore, promote the design and training of practical interpretation algorithms. In general, the efficiency and practicality of the dataset for RS image interpretation can be greatly improved by cooperating with application departments.
   
    {\em Tools for RS image annotation.} 
    Another point worth noting is the necessity of developing professional and open-sourced tools for RS image annotation. A number of popular tools for image annotation have been published, as listed in Table~\ref{tab:AnnotationTools}. These include excellent tools for specific image content interpretation tasks, \eg, object recognition~\cite{LabelImg,VOTT,PolygonRNN++}. Some annotation tools strive to provide diverse annotation modalities, such as polygon, rectangle, circle, ellipse, line, and point~\cite{LabelMe,CVAT,imagetagger,makesenseai,VIA}, serving as universal annotation platforms that are applicable to build image-level labels, the local extent of objects, and semantic information of pixels. Due to the differences in interpretation tasks and application requirements, the most common concerns among annotators are the features and instructions of these tools. The properties of these annotations tools are summarized in Table~\ref{tab:AnnotationTools} and more details can be found in the corresponding reference materials. In general, when building an annotated dataset for RS image content interpretation, the choice of a flexible annotation tool is of great significance for efficiency and quality assurance.
    
    {\em Processing for noisy annotations.}
     The processing of noise annotations and also algorithms tolerant to noise annotations are the common requirements in real application scenarios. In the construction of a RS image dataset, images can be annotated by multiple experts while different annotators possess varying levels of expertise. Besides, the opinions of annotators may conflict with each other because of the personal bias~\cite{khetan2018learning}. Not only that, but RS images with high complexity is hard to be correctly interpreted even for the experts due to the high demand for specialized background and knowledge of RS image interpretation. All of these will inevitably lead to noisy annotations. An intuitive approach to overcome this problem is to remove the noisy annotations by manual cleaning and correction. However, cleansing annotations by the manual way usually results in high costs of time and labor. Thus, how to quickly find out the possible noise annotations in constructed dataset becomes a challenging problem. Faced with this situation, it is valuable to build effective algorithms to model and predict noise annotations for data cleansing and quality improvement. On the other hand, in order to obtain a high-performance algorithm for RS image interpretation, most data-driven methods require a fair amount of data with precise annotations for proper training, particularly the deep learning algorithms. Thus, the effect of noise annotation on the performance of interpretation algorithms is necessary to be explored for better utilization of the annotated dataset~\cite{pelletier2017effect,swan2018good}. Furthermore, it is crucial to consider the existence of annotation noise and develop noise-robust algorithms~\cite{frenay2013classification,li2019learning} to efficiently fade away its negative effects on RS image interpretation.

\section{Conclusions} \label{Conclusions}
RS technology over the past years has made tremendous progress and been providing us a huge amount of RS images for systematic observation of the earth surface. However, the lack of publicly available large-scale RS image datasets with accurate annotation has become a bottle-neck problem to the development of new and intelligent approaches for image interpretation. 

Through a bibliometric analysis, this paper first presents a systematic review of the existing datasets related to the mainstream of RS image interpretation tasks. It reveals that many of the annotated RS image datasets, to some extent, show deficiencies in one or several different aspects, \eg, diversity and scale, that hamper the development of practical interpretation models. Hence, the creation of RS image datasets needs to be paid with more attention, from the annotation process to property control for real applications. Subsequently, we paid efforts to explore the guidances for building the useful dataset for RS image interpretation. It is suggested that the construction of the RS image datasets should be created toward the requirements of practical applications, rather than the interpretation algorithms. The presented guidances formulates a prototype for RS image dataset construction with consideration in efficiency and quality assurance. With the introduced guidances, we created a large-scale RS image dataset for scene classification, \ie, Million-AID, through a semi-automatic annotation strategy. It provides a new idea and approach for the construction of RS image datasets. And the discussion about challenges and perspectives in RS image dataset annotation delivers a new sight for the future work where efforts need to be dedicated for RS image dataset annotation.

In the future, we will devote our endeavor to develop a publicly online evaluation platform for various interpretation datasets and algorithms. We believe that the trend of intelligent interpretation for RS images is unstoppable, and more practical datasets and algorithms oriented to real RS applications will be created in the coming years. It should be encouraged that more datasets and interpretation frameworks be shared within the RS community to advance the prosperity of intelligent interpretation and applications of RS images.

{\small
\bibliographystyle{IEEEtran}
\bibliography{main-ieee}
}

\end{document}